\documentclass[a4paper]{article}
\usepackage{times}
\usepackage{helvet}
\usepackage{fancyhdr}
\usepackage{graphicx}
\usepackage[round,authoryear]{natbib}
\usepackage{subfigure}
\usepackage{nicefrac}
\usepackage{booktabs}
\usepackage[T1]{fontenc}
\usepackage{multirow}
\usepackage{amsmath,amssymb,amsbsy}
\usepackage{textcomp}
\usepackage{pdflscape}
\usepackage[hyphens]{url}

\newtheorem{theorem}{Theorem}
\newtheorem{lemma}[theorem]{Lemma}

\usepackage[accepted]{icml2017}

\usepackage{xr}
\DeclareMathOperator{\E}{E}
\DeclareMathOperator*{\Var}{Var}
\DeclareMathOperator{\sign}{sign}

\DeclareMathOperator*{\argmin}{argmin}
\DeclareMathOperator*{\argmax}{argmax}

\newcommand{\calXp}{\mathcal{X}_\mathrm{P}}
\newcommand{\calXn}{\mathcal{X}_\mathrm{N}}
\newcommand{\calXu}{\mathcal{X}_\mathrm{U}}

\newcommand{\rP}{\mathrm{P}}
\newcommand{\rN}{\mathrm{N}}

\newcommand{\thetap}{\theta_\mathrm{P}}
\newcommand{\thetan}{\theta_\mathrm{N}}

\newcommand{\PN}{\mathrm{PN}}
\newcommand{\NPU}{{\mathrm{N}\textrm{-}\mathrm{PU}}}
\newcommand{\CPU}{{\mathrm{C}\textrm{-}\mathrm{PU}}}
\newcommand{\PU}{\mathrm{PU}}
\newcommand{\NNU}{{\mathrm{N}\textrm{-}\mathrm{NU}}}
\newcommand{\CNU}{{\mathrm{C}\textrm{-}\mathrm{NU}}}
\newcommand{\NU}{\mathrm{NU}}

\newcommand{\PNU}{\mathrm{PNU}}
\newcommand{\NPNPU}{{\mathrm{N}\textrm{-}\mathrm{PNPU}}}
\newcommand{\CPNPU}{{\mathrm{C}\textrm{-}\mathrm{PNPU}}}
\newcommand{\PNPU}{\mathrm{PNPU}}
\newcommand{\NPNNU}{{\mathrm{N}\textrm{-}\mathrm{PNNU}}}
\newcommand{\CPNNU}{{\mathrm{C}\textrm{-}\mathrm{PNNU}}}
\newcommand{\PNNU}{\mathrm{PNNU}}
\newcommand{\NPUNU}{{\mathrm{N}\textrm{-}\mathrm{PUNU}}}
\newcommand{\CPUNU}{{\mathrm{C}\textrm{-}\mathrm{PUNU}}}
\newcommand{\PUNU}{\mathrm{PUNU}}

\newcommand{\ellt}{{\widetilde{\ell}}} 
\newcommand{\ellzo}{{\ell_{0\textrm{-}1}}} 
\newcommand{\ellLin}{{\ell_\mathrm{Lin}}} 
\newcommand{\ellLog}{{\ell_\mathrm{L}}} 
\newcommand{\ellS}{{\ell_\mathrm{S}}} 
\newcommand{\ellH}{{\ell_\mathrm{H}}} 
\newcommand{\ellR}{{\ell_\mathrm{R}}} 
\newcommand{\ellDH}{{\ell_\mathrm{DH}}} 
\newcommand{\ellTS}{{\ell_\mathrm{TS}}} 

\newcommand{\bxp}{\boldsymbol{x}^\mathrm{P}}
\newcommand{\bxn}{\boldsymbol{x}^\mathrm{N}}
\newcommand{\bxu}{\boldsymbol{x}^\mathrm{U}}
\newcommand{\np}{{n_\mathrm{P}}}
\newcommand{\nn}{{n_\mathrm{N}}}
\newcommand{\nun}{{n_\mathrm{U}}}
\newcommand{\nl}{{n_\mathrm{L}}}

\newcommand{\un}{\mathrm{U}}
\newcommand{\sigmap}{\sigma_\mathrm{P}}
\newcommand{\sigman}{\sigma_\mathrm{N}}
\newcommand{\Varp}{\Var\nolimits_\mathrm{P}}
\newcommand{\Varn}{\Var\nolimits_\mathrm{N}}
\newcommand{\Ep}{\E_\mathrm{P}}
\newcommand{\En}{\E_\mathrm{N}}
\newcommand{\Eu}{\E_\mathrm{U}}
\newcommand{\Rp}{R_\mathrm{P}}
\newcommand{\Rn}{R_\mathrm{N}}
\newcommand{\Rup}{R_{\mathrm{U,P}}}
\newcommand{\Run}{R_{\mathrm{U,N}}}
\newcommand{\hRp}{\widehat{R}_{\mathrm{P}}}
\newcommand{\hRn}{\widehat{R}_{\mathrm{N}}}
\newcommand{\hRup}{\widehat{R}_{\mathrm{U,P}}}
\newcommand{\hRun}{\widehat{R}_{\mathrm{U,N}}}
\newcommand{\psip}{\psi_\mathrm{P}}
\newcommand{\psin}{\psi_\mathrm{N}}
\newcommand{\const}{\mathrm{Const}}

\newcommand{\Rh}{\widehat{R}}
\newcommand{\RpL}{R_\mathrm{P}^\mathrm{L}}
\newcommand{\RnL}{R_\mathrm{N}^\mathrm{L}}
\newcommand{\RpC}{R_\mathrm{P}^\mathrm{C}}
\newcommand{\RnC}{R_\mathrm{N}^\mathrm{C}}

\newcommand{\bA}{{\boldsymbol{A}}}

\newcommand{\bx}{{\boldsymbol{x}}}

\newcommand{\bD}{{\boldsymbol{D}}}

\newcommand{\bk}{{\boldsymbol{k}}}
\newcommand{\bK}{{\boldsymbol{K}}}
\newcommand{\bL}{{\boldsymbol{L}}}

\newcommand{\bt}{{\boldsymbol{t}}}

\newcommand{\by}{{\boldsymbol{y}}}

\newcommand{\bw}{{\boldsymbol{w}}}
\newcommand{\bW}{{\boldsymbol{W}}}

\newcommand{\balpha}{{\boldsymbol{\alpha}}}

\newcommand{\bphi}{{\boldsymbol{\phi}}}

\newcommand{\bone}{{\boldsymbol{1}}}

\newcommand{\calA}{\mathcal{A}}
\newcommand{\calB}{\mathcal{B}}
\newcommand{\calC}{\mathcal{C}}
\newcommand{\calG}{\mathcal{G}}

\newcommand{\calL}{\mathcal{L}}
\newcommand{\calM}{\mathcal{M}}
\newcommand{\calO}{\mathcal{O}}
\newcommand{\calU}{\mathcal{U}}
\newcommand{\calX}{\mathcal{X}}
\newcommand{\calY}{\mathcal{Y}}

\newcommand{\nhalf}{{-\frac{1}{2}}}

\newcommand{\tr}{\mathrm{tr}}

\newcommand{\diag}{\mathrm{diag}}

\newcommand{\from}{\colon}

\newcommand{\ysvm}{y^{svm}}
\newcommand{\yh}{\widehat{y}}
\newcommand{\yb}{\bar{y}}
\newcommand{\byh}{\boldsymbol{\widehat{y}}}
\newcommand{\byb}{\boldsymbol{\bar{y}}}
\newcommand{\bysvm}{\boldsymbol{y}^\mathrm{svm}}

\newcommand{\ED}{\textrm{ED}}
\newcommand{\bxch}{{\boldsymbol{\check{x}}}}

\icmltitlerunning{Semi-Supervised Classification 
	Based on Classification from Positive and Unlabeled Data}

\begin{document}
\sloppy

\twocolumn[
\icmltitle{Semi-Supervised Classification \\ 
	Based on Classification from Positive and Unlabeled Data}

\begin{icmlauthorlist}
\icmlauthor{Tomoya Sakai}{UTokyo,RIKEN}
\hspace{2mm}
\icmlauthor{Marthinus Christoffel du Plessis}{}
\hspace{2mm}
\icmlauthor{Gang Niu}{UTokyo}
\hspace{2mm}
\icmlauthor{Masashi Sugiyama}{RIKEN,UTokyo}
\end{icmlauthorlist}

\icmlaffiliation{UTokyo}{The University of Tokyo, Japan}
\icmlaffiliation{RIKEN}{RIKEN, Japan}

\icmlcorrespondingauthor{Tomoya Sakai}{sakai@ms.k.u-tokyo.ac.jp}

\icmlkeywords{
classification,
semi-supervised classification,
classification from positive and unlabeled data
}

\vskip 0.3in
]

\printAffiliationsAndNotice{}

\begin{abstract}%
Most of the semi-supervised classification methods
developed so far use unlabeled data for regularization purposes
under particular distributional assumptions such as
the cluster assumption.
In contrast, recently developed methods of
\emph{classification from positive and unlabeled data} (PU classification)
use unlabeled data for risk evaluation,
i.e., label information is directly extracted from unlabeled data.
In this paper, we extend PU classification to also incorporate negative data
and propose a novel semi-supervised classification approach.
We establish generalization error bounds for our novel methods
and show that the bounds decrease with respect to the number
of unlabeled data \emph{without} the distributional assumptions
that are required in existing semi-supervised classification methods.
Through experiments, we demonstrate the usefulness of the proposed methods.
\end{abstract}

\section{Introduction}
\label{sec:introduction}
Collecting a large amount of labeled data is
a critical bottleneck in real-world machine learning applications
due to the laborious manual annotation.
In contrast, unlabeled data can often be collected automatically and abundantly,
e.g., by a web crawler.
This has led to the development of 
various semi-supervised classification algorithms over the past decades.

To leverage unlabeled data in training, 
most of the existing semi-supervised classification methods 
rely on particular assumptions on the data distribution
\citep{book:Chapelle+etal:2006}.
For example, the \emph{manifold assumption} supposes that 
samples are distributed on a low-dimensional manifold 
in the data space \citep{JMLR:Belkin+etal:2006}.
In the existing framework, such a distributional assumption is 
encoded as a regularizer for training a classifier
and \emph{biases} the classifier toward a better one 
under the assumption.
However, if such a distributional assumption contradicts the data distribution,
the bias behaves adversely, and the performance of the obtained classifier
becomes worse than the one obtained with supervised classification
\citep{ICML:Cozman+etal:2003,ICML:Sokolovska+etal:2008,PAMI:Li+Zhou:2015,PR:Krijthe:2017}.

Recently, 
\emph{classification from positive and unlabeled data} 
(PU classification)
has been gathering growing attention 
\citep{KDD:Elkan+Noto:2008,NIPS:duPlessis+etal:2014,ICML:duPlessis+etal:2015,NIPS:Jain+etal:2016},
which trains a classifier only from positive and unlabeled data 
without negative data.
In PU classification, the \emph{unbiased} risk estimators proposed
in \citet{NIPS:duPlessis+etal:2014,ICML:duPlessis+etal:2015} 
utilize unlabeled data for \emph{risk evaluation}, 
implying that label information is directly extracted from unlabeled data
without restrictive distributional assumptions, 
unlike existing semi-supervised classification methods
that utilize unlabeled data for \emph{regularization}.
Furthermore, 
theoretical analysis \citep{NIPS:Niu+etal:2016}
showed that PU classification (or its counterpart, \emph{NU classification},
classification from negative and unlabeled data) is likely to outperform 
classification from positive and negative data (\emph{PN classification}, 
i.e., ordinary supervised classification) depending on 
the number of positive, negative, and unlabeled samples.
It is thus naturally expected that combining PN, PU, and NU classification 
can be a promising approach to semi-supervised classification 
without restrictive distributional assumptions. 

In this paper, we propose a novel semi-supervised classification approach
by considering convex combinations of the risk functions of PN, PU, 
and NU classification.
Without any distributional assumption,   
we theoretically show that 
the confidence term of the generalization error bounds
decreases at the optimal parametric rate with respect to 
the number of positive, negative, and unlabeled samples,
and the variance of the proposed risk estimator is 
almost always smaller than the plain PN risk function
given an infinite number of unlabeled samples.
Through experiments, we analyze the behavior of 
the proposed approach 
and demonstrate the usefulness of the proposed semi-supervised 
classification methods.


\section{Background}
\label{sec:background}
In this section, we first introduce the notation commonly used in 
this paper and review the formulations of PN, PU, and NU classification.

\subsection{Notation}
Let random variables $\bx\in\mathbb{R}^d$ and $y\in\{+1,-1\}$ be equipped with 
probability density $p(\bx,y)$, where $d$ is a positive integer. 
Let us consider a binary classification problem from $\bx$ to $y$,
given three sets of samples called the \emph{positive}~(P), \emph{negative}~(N),
and \emph{unlabeled}~(U) data: 
\begin{align*}
\calX_\mathrm{P} &:= \{\bxp_i\}^{\np}_{i=1} \stackrel{\mathrm{i.i.d.}}{\sim}
 p_\mathrm{P}(\bx):= p(\bx \mid y=+1) , \\ 
\calX_\mathrm{N} &:= \{\bxn_i\}^{\nn}_{i=1} \stackrel{\mathrm{i.i.d.}}{\sim}
 p_\mathrm{N}(\bx):=p(\bx \mid y=-1) , \\ 
\calX_\un &:= \{\bx^\un_i\}^{\nun}_{i=1} \stackrel{\mathrm{i.i.d.}}{\sim}
 p(\bx):=\thetap p_\mathrm{P}(\bx)+\thetan p_\mathrm{N}(\bx),
\end{align*}
where 
\begin{align*}
  \thetap:=p(y=+1),~~~~
  \thetan:=p(y=-1)
\end{align*}
are the class-prior probabilities
for the positive and negative classes such that $\thetap+\thetan=1$.

Let $g\from\mathbb{R}^d\to\mathbb{R}$ be an arbitrary real-valued
decision function for binary classification, and
classification is performed based on its sign.
Let $\ell\from\mathbb{R}\to\mathbb{R}$ be a loss function
such that $\ell(m)$ generally takes a small value for large margin $m=yg(\bx)$.
Let $\Rp(g)$, $\Rn(g)$, $\Rup(g)$, and $\Run(g)$ be
the risks of classifier $g$ under loss $\ell$:
\begin{align*}
  \Rp(g)&:=\Ep[\ell(g(\bx))],& \Rn(g)&:=\En[\ell(-g(\bx))],\\
  \Rup(g)&:=\E_\un[\ell(g(\bx))], &\Run(g)&:=\E_\un[\ell(-g(\bx))],
\end{align*}
where $\Ep$, $\En$, and $\E_\un$ denote the expectations
over $p_\mathrm{P}(\bx)$, $p_\mathrm{N}(\bx)$, and $p(\bx)$, respectively.
Since we do not have any samples from $p(\bx,y)$, 
the true risk $R(g)=\E_{p(\bx,y)}[\ell(yg(\bx))]$, which we want to minimize, 
should be recovered without using $p(\bx,y)$ as shown below.



\subsection{PN Classification}
In standard supervised classification (PN classification), 
we have both positive and negative data, i.e., fully labeled data.
The goal of PN classification is to train a classifier using 
labeled data. 

The risk in PN classification (the PN risk) is defined as 
\begin{align}
R_\PN(g)&:=\thetap\Ep[\ell(g(\bx))] + \thetan\En[\ell(-g(\bx))] \notag \\ 
  &\phantom{:}= \thetap \Rp(g) + \thetan \Rn(g) ,
\label{eq:risk-pn}
\end{align}
which is equal to $R(g)$, but $p(\bx,y)$ is not included.
If we use the hinge loss function $\ellH(m):=\max(0,1-m)$, 
the PN risk coincides with 
the risk of the support vector machine \citep{book:Vapnik:1995}.	


\subsection{PU Classification}
In PU classification,
we do not have labeled data for the negative class,
but we can use unlabeled data drawn from marginal density $p(\bx)$.
The goal of PU classification is to train a classifier using only 
positive and unlabeled data. 
The basic approach to PU classification is to 
discriminate P and U data \citep{KDD:Elkan+Noto:2008}.
However, naively classifying P and U data causes a bias.

To address this problem, 
\citet{NIPS:duPlessis+etal:2014,ICML:duPlessis+etal:2015} 
proposed a risk equivalent to the PN risk but 
where $p_\mathrm{N}(\bx)$ is not included.
The key idea is to utilize unlabeled data to evaluate
the risk for negative samples in the PN risk.
Replacing the second term in Eq. \eqref{eq:risk-pn} with\footnote{
The equation comes from the definition of the marginal density 
$p(\bx)=\thetap p_\mathrm{P}(\bx) + \thetan p_\mathrm{N}(\bx)$.}
\begin{align*}
\thetan\En[\ell(-g(\bx))]
&=\Eu[\ell(-g(\bx))] - \thetap\Ep[\ell(-g(\bx))], 
\end{align*}
we obtain the risk in PU classification (the PU risk) as
\begin{align}
R_\PU(g)&:=\thetap\Ep[\ellt(g(\bx))] + \Eu[\ell(-g(\bx))] \notag \\
&\phantom{:}=\thetap\RpC(g) + \Run(g) ,
\label{eq:risk-pu}
\end{align}
where $\RpC(g):=\Ep[\ellt(g(\bx))]$ and
$\ellt(m)=\ell(m)-\ell(-m)$ is a composite loss function.

\paragraph{Non-Convex Approach:}
If the loss function satisfies
\begin{align}
\ell(m)+\ell(-m)=1,
\label{eq:cond-ncvx-pu}
\end{align}
the composite loss function becomes $\ellt(m)=2\ell(m)-1$.
We thus obtain the \emph{non-convex} PU risk as
\begin{align}
  R_\NPU(g):=2\thetap \Rp(g) + \Run(g) - \thetap .
  \label{eq:risk-ncvx-pu}
\end{align}
This formulation can be seen as cost-sensitive classification of 
P and U data with weight $2\thetap$ \citep{NIPS:duPlessis+etal:2014}.

The ramp loss used in the robust support vector machine \citep{ICML:Collobert+etal:2006},
\begin{align}
\ellR(m):=\frac{1}{2}\max(0, \min(2, 1-m)),
\label{ramp-loss}
\end{align}
satisfies the condition \eqref{eq:cond-ncvx-pu}.
However, the use of the ramp loss
(and any other losses that satisfy the condition \eqref{eq:cond-ncvx-pu})
yields a non-convex optimization problem,
which may be solved locally by
the \emph{concave-convex procedure} (CCCP)
\citep{NIPS:Yuille+Rangarajan:2002,ICML:Collobert+etal:2006,NIPS:duPlessis+etal:2014}.


\paragraph{Convex Approach:}
If a convex surrogate loss function satisfies 
\begin{align}
\label{eq:cond-cvx-pu}
\ell(m)-\ell(-m)=-m ,
\end{align}
the composite loss function becomes a linear function
$\ellt(m)=-m$
\citep[see Table $1$ in ][]{ICML:duPlessis+etal:2015}.
We thus obtain the \emph{convex} PU risk as
\begin{align*}
R_\CPU(g):=\thetap \RpL(g) + \Run(g) , 
\end{align*}
where $\RpL(g):=\Ep[-g(\bx)]$ is the risk with the linear loss 
$\ellLin(m):=-m$.
This formulation yields the convex optimization problem
that can be solved efficiently.

\subsection{NU Classification}
As a mirror of PU classification, we can consider NU classification.
The risk in NU classification (the NU risk) is given by
\begin{align*}
R_\NU(g)&:=\thetan\En[\ellt(-g(\bx))] + \Eu[\ell(g(\bx))] \\ 
&\phantom{:}=\thetan\RnC(g) + \Rup(g) ,
\end{align*}
where $\RnC(g):=\En[\ellt(-g(\bx))]$ is 
the risk function with the composite loss.
Similarly to PU classification, 
the non-convex and convex NU risks are
expressed as
\begin{align}
  R_\NNU(g)&:=2\thetan \Rn(g) + \Rup(g) - \thetan, 
  \label{eq:risk-ncvx-nu} \\
  R_\CNU(g)&:=\thetan \RnL(g)+ \Rup(g) ,
\end{align}
where $\RnL(g):=\En[g(\bx)]$ is the risk with the linear loss.

\section{Semi-Supervised Classification Based on PN, PU, and NU Classification}
\label{sec:proposed}
In this section, we propose semi-supervised classification methods
based on PN, PU, and NU classification.

\subsection{PUNU Classification}
A naive idea to build a semi-supervised classifier
is to combine the PU and NU risks.
For $\gamma\in[0,1]$, let us consider a linear combination of
the PU and NU risks:
\begin{align*}
R_\PUNU^\gamma(g)&:=(1-\gamma) R_\PU(g) + \gamma R_\NU(g).
\end{align*}
We refer to this combined method as 
\emph{PUNU classification}.

If we use a loss function satisfying the condition \eqref{eq:cond-ncvx-pu},
the non-convex PUNU risk $R_\NPUNU^\gamma(g)$ can be expressed as
\begin{align*}
R_\NPUNU^\gamma(g) 
&=2(1-\gamma)\thetap\Rp(g)
	+ 2\gamma\thetan\Rn(g) \\
&\phantom{=}+ \E_\un[(1-\gamma)\ell(-g(\bx)) + \gamma\ell(g(\bx))] \\
&\phantom{=}   
	-(1-\gamma)\thetap-\gamma\thetan .
\end{align*}
Here, $R_\NPUNU^{1/2}(g)$ agrees with 
$R_\PN(g)$ due to the condition \eqref{eq:cond-ncvx-pu}.
Thus, when $\gamma=1/2$, 
PUNU classification is reduced to ordinary PN classification.

On the other hand, $\gamma=1/2$ is still effective 
when the condition \eqref{eq:cond-cvx-pu} is satisfied.
Its risk $R_\CPUNU^\gamma(g)$ can be expressed as
\begin{align*}
  R_\CPUNU^\gamma(g)
&=(1-\gamma)\thetap \RpL(g)
+ \gamma\thetan \RnL(g) \\
&\phantom{=}
+ \E_\un[(1-\gamma)\ell(g(\bx))+\gamma\ell(-g(\bx)) ].
\end{align*}
Here, $(1-\gamma)\ell(g(\bx)) + \gamma\ell(-g(\bx))$
can be regarded as a loss function for
unlabeled samples with weight $\gamma$.

When $\gamma=1/2$, unlabeled samples incur the same loss
for the positive and negative classes.
On the other hand, when $0<\gamma<1/2$,
a smaller loss is incurred for the negative class
than the positive class.
Thus, unlabeled samples tend to be classified into the negative class.
The opposite is true when $1/2<\gamma<1$.

\subsection{PNU Classification}
Another possibility of using PU and NU classification 
in semi-supervised classification is to combine 
the PN and PU/NU risks.
For $\gamma\in[0,1]$, let us consider linear combinations of
the PN and PU/NU risks:
\begin{align*}
R_\PNPU^\gamma(g) &:= (1-\gamma) R_\PN(g) + \gamma R_\PU(g),\\
R_\PNNU^\gamma(g) &:= (1-\gamma) R_\PN(g) + \gamma R_\NU(g).
\end{align*}
In practice, we combine PNPU and PNNU classification
and adaptively choose one of them
with a new trade-off parameter $\eta\in[-1,1]$ as
\begin{align*}
  R_\PNU^\eta(g)&:=
  \begin{cases}
    R_\PNPU^\eta(g) &(\eta \geq 0), \\
    R_\PNNU^{-\eta}(g) & (\eta < 0).
  \end{cases}
\end{align*}
We refer to the combined method as \emph{PNU classification}.
Clearly, PNU classification with $\eta=-1,0,+1$ corresponds to 
NU, PN, and PU classification. 
As $\eta$ gets large/small,
the effect of the positive/negative classes is more emphasized.

In the theoretical analyses in Section~\ref{sec:theory},
we denote the combinations of the PN risk with the non-convex PU/NU risks by
$R_\NPNPU^\gamma$ and $R_\NPNNU^\gamma$,
and that with the convex PU/NU risks by
$R_\CPNPU^\gamma$ and $R_\CPNNU^\gamma$.

\subsection{Practical Implementation}
We have so far only considered the true risks $R$
(with respect to the expectations over true data distributions).
When a classifier is trained from samples in practice,
we use the empirical risks $\widehat{R}$
where the expectations are replaced with corresponding sample averages.

More specifically, 
in the theoretical analysis in Section~\ref{sec:theory}
and experiments in Section~\ref{sec:experiments},
we use a linear-in-parameter model given by
\[
g(\bx)=\sum^b_{j=1}w_j\phi_j(\bx)=\bw^\top\bphi(\bx),
\]
where $^\top$ denotes the transpose,
$b$ is the number of basis functions,
$\bw=(w_1, \dots, w_b)^\top$ is a parameter vector,
and $\bphi(\bx)=(\phi_1(\bx), \dots, \phi_b(\bx))^\top$
is a basis function vector.
The parameter vector $\bw$ is learned in order to minimize
the $\ell_2$-regularized empirical risk:
\begin{align*}
  \min_\bw \Rh(g)+\lambda\bw^\top\bw,
\end{align*}
where $\lambda\ge0$ is the regularization parameter.

\section{Theoretical Analyses}
\label{sec:theory}%
In this section, we theoretically analyze the behavior of 
the empirical versions of the proposed semi-supervised
classification methods.
We first derive generalization error bounds 
and then discuss variance reduction.
Finally, we discuss which approach, 
PUNU or PNU classification, is more promising.
All proofs can be found in Appendix~\ref{sec:proofs}.

\subsection{Generalization Error Bounds}
\label{sec:theory-gen-err}%

Let $\calG$ be a function class of bounded hyperplanes:
\begin{equation*}
\calG = \{g(\bx)=\langle \bw,\bphi(\bx) \rangle 
\mid \|\bw\|\le C_w, \|\bphi(\bx)\|\le C_\phi \},
\end{equation*}
where $C_w$ and $C_\phi$ are certain positive constants. 
Since $\ell_2$-regularization is always included, 
we can naturally assume that the empirical risk minimizer $g$ belongs to a certain $\calG$. 
Denote by $\ellzo(m)=(1-\sign(m))/2$ the \emph{zero-one loss} 
and $I(g)=\E_{p(\bx,y)}[\ellzo(yg(\bx))]$ the risk of $g$ for binary
classification, i.e., the generalization error of $g$. 
In the following, we study upper bounds of $I(g)$ holding 
uniformly for all $g\in\calG$.
We respectively focus on the \emph{(scaled) ramp and squared losses} for the
non-convex and convex methods due to limited space.
Similar results can be obtained with a little more effort if other eligible losses are used. 
For convenience, we define a function as
\begin{align*}
\chi(c_\mathrm{P},c_\mathrm{N},c_\un)
=c_\mathrm{P}\thetap/\sqrt{\np}
+c_\mathrm{N}\thetan/\sqrt{\nn}
+c_\un/\sqrt{\nun}.
\end{align*}

\paragraph{Non-Convex Methods:}
A key observation is that $\ellzo(m)\le2\ellR(m)$, and consequently $I(g)\le2R(g)$. 
Note that by definition we have
\begin{equation*}
R_\NPUNU^\gamma(g)=R_\NPNPU^\gamma(g)=R_\NPNNU^\gamma(g)=R(g) .
\end{equation*}
The theorem below can be proven using the Rademacher analysis \citep[see, for example,][]{book:Mohri+etal:2012,book:Ledoux+Talagrand:1991}.

\begin{theorem}
  \label{thm:gen-err-rl}%
  Let $\ellR(m)$ be the loss for defining the empirical risks. For any $\delta>0$, 
  the following inequalities hold separately with probability at least $1-\delta$ 
  for all $g\in\calG$:
  \begin{align*}
  I(g) &\le 2\Rh_\NPUNU^\gamma(g)+C_{w,\phi,\delta}\cdot\chi(2-2\gamma,2\gamma,|2\gamma-1|),\\
  I(g) &\le 2\Rh_\NPNPU^\gamma(g)+C_{w,\phi,\delta}\cdot\chi(1+\gamma,1-\gamma,\gamma),\\
  I(g) &\le 2\Rh_\NPNNU^\gamma(g)+C_{w,\phi,\delta}\cdot\chi(1-\gamma,1+\gamma,\gamma),
  \end{align*} 
  where $C_{w,\phi,\delta}=2C_wC_\phi +\sqrt{2\ln(3/\delta)}$.
\end{theorem}

Theorem~\ref{thm:gen-err-rl} guarantees that when $\ellR(m)$ is used, 
$I(g)$ can be bounded from above by two times the empirical risks, i.e., 
$2\Rh_\NPUNU^\gamma(g)$, $2\Rh_\NPNPU^\gamma(g)$, and $2\Rh_\NPNNU^\gamma(g)$,
plus the corresponding confidence terms of order
\begin{equation*}
\calO_p(1/\sqrt{\np}+1/\sqrt{\nn}+1/\sqrt{\nun}).
\end{equation*}
Since $\np$, $\nn$, and $\nun$ can increase independently, 
this is already the optimal convergence rate without any additional assumption
\citep{book:Vapnik:1998,TIT:Mendelson:2008}.

\paragraph{Convex Methods:}
Analogously, we have $\ellzo(m)\le4\ellS(m)$ for the squared loss. 
However, it is too loose when $|m|\gg0$. Fortunately, we do not have
to use $\ellS(m)$ if we work on the generalization error rather than the estimation error. 
To this end, we define the \emph{truncated (scaled) squared loss} $\ellTS(m)$ as
\begin{equation*}
\ellTS(m)=
\begin{cases}
\ellS(m) & 0<m\le1,\\
\ellzo(m)/4 & \textrm{otherwise},
\end{cases}
\end{equation*}
so that $\ellzo(m)\le4\ellTS(m)$ is much tighter. 
For $\ellTS(m)$, $R_\CPU(g)$ and $R_\CNU(g)$ need to be redefined as follows 
\citep[see][]{ICML:duPlessis+etal:2015}:
\begin{align*}
R_\CPU(g)&:=\thetap\Rp'(g)+\Run(g),\\
R_\CNU(g)&:=\thetan\Rn'(g)+\Rup(g),
\end{align*}
where $\Rp'(g)$ and $\Rn'(g)$ are simply $\Rp(g)$ and $\Rn(g)$ w.r.t.\ 
the composite loss $\ellt_\mathrm{TS}(m)=\ellTS(m)-\ellTS(-m)$. 
The condition $\ellt_\mathrm{TS}(m)\neq-m$ means 
the loss of convexity, but the equivalence is not lost;
indeed, we still have
\begin{equation*}
R_\CPUNU^\gamma(g)=R_\CPNPU^\gamma(g)=R_\CPNNU^\gamma(g)=R(g).
\end{equation*}

\begin{theorem}
  \label{thm:gen-err-sl}%
  Let $\ellTS(m)$ be the loss for defining the empirical risks (where $R_\CPU(g)$ and $R_\CNU(g)$ are redefined).
  For any $\delta>0$, the following inequalities hold separately with probability at least $1-\delta$ for all
$g\in\calG$:
  \begin{align*}
  I(g) &\le 4\Rh_\CPUNU^\gamma(g)+C'_{w,\phi,\delta}\cdot\chi(1-\gamma,\gamma,1),\\
  I(g) &\le 4\Rh_\CPNPU^\gamma(g)+C'_{w,\phi,\delta}\cdot\chi(1,1-\gamma,\gamma),\\
  I(g) &\le 4\Rh_\CPNNU^\gamma(g)+C'_{w,\phi,\delta}\cdot\chi(1-\gamma,1,\gamma),
  \end{align*} 
  where $C'_{w,\phi,\delta}=4C_wC_\phi +\sqrt{2\ln(4/\delta)}$.
\end{theorem}

Theorem~\ref{thm:gen-err-sl} ensures that when $\ellTS(m)$ is used 
(for evaluating the empirical risks rather than learning the empirical risk minimizers), 
$I(g)$ can be bounded from above by four times the empirical risks plus confidence terms 
in the optimal parametric rate. 
As $\ellTS(m)\le\ellS(m)$, Theorem~\ref{thm:gen-err-sl} is valid 
(but weaker) if all empirical risks are w.r.t.\ $\ellS(m)$.

\subsection{Variance Reduction}
\label{sec:theory-var}%

Our empirical risk estimators proposed in Section~\ref{sec:proposed} 
are all unbiased.
The next question is whether their variance can be smaller than that of $\Rh_\PN(g)$, 
i.e., whether $\calX_\un$ can help reduce the variance in estimating $R(g)$.
To answer this question, pick any $g$ of interest. 
For simplicity, we assume that $\nun\rightarrow\infty$, 
to illustrate the maximum variance reduction that could be achieved. 
Due to limited space, we only focus on the non-convex methods.

Similarly to $\Rp(g)$ and $\Rn(g)$, let $\sigmap^2(g)$ and $\sigman^2(g)$ 
be the corresponding variance:
\begin{align*}
\sigmap^2(g):=\Varp[\ell(g(\bx))],\quad
\sigman^2(g):=\Varn[\ell(-g(\bx))],
\end{align*}
where $\Varp$ and $\Varn$ denote the variance over $p_\mathrm{P}(\bx)$ and $p_\mathrm{N}(\bx)$.
Moreover, denote by $\psip=\thetap^2\sigmap^2(g)/\np$ and
$\psin=\thetan^2\sigman^2(g)/\nn$ for short,
and let $\Var$ be the variance over 
$p_\mathrm{P}(\bxp_1)\cdots p_\mathrm{P}(\bxp_\np)\cdot
p_\mathrm{N}(\bxn_1)\cdots p_\mathrm{N}(\bxn_\nn)\cdot
p(\bxu_1)\cdots p(\bxu_\nun)$.

\begin{theorem}
  \label{thm:var-ncvx-punu}%
  Assume $\nun\rightarrow\infty$. For any fixed $g$, let
  \begin{align} 
  \gamma_\NPUNU=\argmin_\gamma\Var[\Rh_\NPUNU^\gamma(g)]=\frac{\psip}{\psip+\psin}.
  \label{eq:gam-npunu}
  \end{align}
  Then, we have $\gamma_\NPUNU\in[0,1]$.
  Further, $\Var[\Rh_\NPUNU^\gamma(g)]<\Var[\Rh_\PN(g)]$ for all
  $\gamma\in(2\gamma_\NPUNU-1/2,1/2)$
  if $\psip<\psin$, or for all $\gamma\in(1/2,2\gamma_\NPUNU-1/2)$ 
  if $\psip>\psin$.%
  \footnote{Being fixed means $g$ is determined before seeing the data 
  for evaluating the empirical risk. 
  For example, if $g$ is trained by some learning method, 
  and the empirical risk is subsequently evaluated on the validation/test data, 
  $g$ is regarded as fixed in the evaluation.}
\end{theorem}

Theorem~\ref{thm:var-ncvx-punu} guarantees that the variance 
is always reduced by $\Rh_\NPUNU^\gamma(g)$ 
if $\gamma$ is close to $\gamma_\NPUNU$, 
which is optimal for variance reduction.
The interval of such good $\gamma$ values 
has the length $\min\{|\psip-\psin|/(\psip+\psin),1/2\}$. 
In particular, if $3\psip\le\psin$ or $\psip\ge3\psin$, the length is $1/2$.

\begin{theorem}
  \label{thm:var-ncvx-pnu}%
  Assume $\nun\rightarrow\infty$. For any fixed $g$, let
  \begin{align}
  \gamma_\NPNPU&\!\!=\argmin_\gamma\Var[\Rh_\NPNPU^\gamma(g)]\!
  =\frac{\psin-\psip}{\psip+\psin},
  \label{eq:gam-npnpu} \\
  \gamma_\NPNNU&\!\!=\argmin_\gamma\Var[\Rh_\NPNNU^\gamma(g)]\!
  =\frac{\psip-\psin}{\psip+\psin}.
  \label{eq:gam-npnnu} 
  \end{align}
  Then, we have $\gamma_\NPNPU\in[0,1]$ if $\psip\le\psin$ or $\gamma_\NPNNU\in[0,1]$ if $\psip\ge\psin$.
  Additionally, $\Var[\Rh_\NPNPU^\gamma(g)]<\Var[\Rh_\PN(g)]$ for all $\gamma\in(0,2\gamma_\NPNPU)$ 
  if $\psip<\psin$, or $\Var[\Rh_\NPNNU^\gamma(g)]<\Var[\Rh_\PN(g)]$ for all $\gamma\in(0,2\gamma_\NPNNU)$ if $\psip>\psin$.
\end{theorem}

Theorem~\ref{thm:var-ncvx-pnu} implies that the variance of $\Rh_\PN(g)$ 
is reduced by either $\Rh_\NPNPU^\gamma(g)$ 
if $\psip\le\psin$ or $\Rh_\NPNNU^\gamma(g)$ if $\psip\ge\psin$, 
where $\gamma$ should be close to $\gamma_\NPNPU$ or $\gamma_\NPNNU$. 
The range of such good $\gamma$ values is of length
$\min\{2|\psip-\psin|/(\psip+\psin),1\}$.
In particular, if $3\psip\le\psin$, $\Rh_\NPNPU^\gamma(g)$ 
given any $\gamma\in(0,1)$ can reduce the variance, 
and if $\psip\ge3\psin$, $\Rh_\NPNNU^\gamma(g)$ given any $\gamma\in(0,1)$ 
can reduce the variance.

As a corollary of Theorems~\ref{thm:var-ncvx-punu} and \ref{thm:var-ncvx-pnu}, 
the minimum variance achievable by $\Rh_\NPUNU^\gamma(g)$, 
$\Rh_\NPNPU^\gamma(g)$, and $\Rh_\NPNNU^\gamma(g)$ 
at their optimal $\gamma_\NPUNU$, $\gamma_\NPNPU$, and $\gamma_\NPNNU$ 
is exactly the same, namely, $4\psip\psin/(\psip+\psin)$. 
Nevertheless, $\Rh_\NPNPU^\gamma(g)$ and $\Rh_\NPNNU^\gamma(g)$ 
have a much wider range of nice $\gamma$ values 
than $\Rh_\NPUNU^\gamma(g)$.

If we further assume that $\sigmap(g)=\sigman(g)$, 
the condition in Theorems~\ref{thm:var-ncvx-punu} and \ref{thm:var-ncvx-pnu} 
as to 	whether $\psip\le\psin$ or $\psip\ge\psin$ will be independent of $g$. 
Also, it will coincide with the condition in \emph{Theorem~7} in \citet{NIPS:Niu+etal:2016} 
where the minimizers of $\Rh_\PN(g)$, $\Rh_\PU(g)$ and $\Rh_\NU(g)$ are compared.

A final remark is that learning is uninvolved in
Theorems~\ref{thm:var-ncvx-punu} and \ref{thm:var-ncvx-pnu}, 
such that $\ell(m)$ can be any loss that satisfies $\ell(m)+\ell(-m)=1$, 
and $g$ can be any fixed decision function. 
For instance, we may adopt $\ellzo(m)$ and pick some $g$ resulted from 
some other learning methods. 
As a consequence, the variance of $\widehat{I}_\PN(g)$ over the validation data can be reduced, 
and then the cross-validation should be more stable, given that $\nun$ is sufficiently large. 
Therefore, even without being minimized, our proposed risk estimators 
are themselves of practical importance.

\subsection{PUNU vs.~PNU Classification}
\label{sec:theory-punu-pnu}
We discuss here which approach,   
PUNU or PNU classification, is more promising
according to state-of-the-art theoretical comparisons
\citep{NIPS:Niu+etal:2016}, which are based on estimation error bounds.

Let $\widehat{g}_\PN$, $\widehat{g}_\PU$, and $\widehat{g}_\NU$
be the minimizers of $\Rh_\PN(g)$, $\Rh_\PU(g)$, and $\Rh_\NU(g)$, 
respectively.
Let $\alpha_{\PU,\PN}:=(\thetap/\sqrt{\np}+1/\sqrt{\nun})
/(\thetan/\sqrt{\nn})$
and
$\alpha_{\NU,\PN}:=(\thetan/\sqrt{\nn}+1/\sqrt{\nun})
/(\thetap/\sqrt{\np})$.
The finite-sample comparisons
state that 
if $\alpha_{\PU,\PN}>1$ ($\alpha_{\NU,\PN}>1$),
PN classification is more promising than PU (NU) classification,
i.e., $R(\widehat{g}_\PN)<R(\widehat{g}_\PU)$
($R(\widehat{g}_\PN)<R(\widehat{g}_\NU)$);
otherwise PU (NU) classification is more promising than PN classification
\citep[cf. Section $3.2$ in][]{NIPS:Niu+etal:2016}.

Suppose that $\nun$ is not sufficiently large against $\np$ and $\nn$.
According to the finite-sample comparisons,
PN classification is most promising, and either PU or NU classification 
is the second best, i.e.,
$R(\widehat{g}_\PN)<R(\widehat{g}_\PU)<R(\widehat{g}_\NU)$ 
or 
$R(\widehat{g}_\PN)<R(\widehat{g}_\NU)<R(\widehat{g}_\PU)$.
On the other hand, if $\nun$ is sufficiently large
($\nun\to\infty$, which is faster than $\np,\nn\to\infty$),
we have the asymptotic comparisons:
$\alpha_{\PU,\PN}^\ast=\lim_{\np,\nn,\nun\to\infty}\alpha_{\PU,\PN}$,
$\alpha_{\NU,\PN}^\ast=\lim_{\np,\nn,\nun\to\infty}\alpha_{\NU,\PN}$,
and $\alpha_{\PU,\PN}^\ast\cdot \alpha_{\NU,\PN}^\ast=1$.
From the last equation, 
if $\alpha_{\PU,\PN}^\ast<1$, then $\alpha_{\NU,\PN}^\ast>1$, 
implying that PU (PN) classification is more promising than PN
(NU) classification, i.e., 
$R(\widehat{g}_\PU)<R(\widehat{g}_\PN)<R(\widehat{g}_\NU)$. 
Similarly,
when $\alpha_{\PU,\PN}^\ast>1$ and $\alpha_{\NU,\PN}^\ast<1$, 
$R(\widehat{g}_\NU)<R(\widehat{g}_\PN)<R(\widehat{g}_\PU)$
\citep[cf. Section $3.3$ in][]{NIPS:Niu+etal:2016}.



In real-world applications,
since we do not know whether the number of unlabeled
samples is sufficiently large or not, 
a practical approach is to combine the best methods
in both the finite-sample and asymptotic cases.
PNU classification is the combination of the best methods 
in both cases, but PUNU classification is not.
In addition, PUNU classification includes 
the worst one in its combination in both cases.
From this viewpoint, PNU classification
would be more promising than PUNU classification,
as demonstrated in the experiments shown in the next section.



\section{Experiments}
\label{sec:experiments}
In this section, we first numerically analyze the proposed approach 
and then compare the proposed semi-supervised classification methods 
against existing methods.
All experiments were carried out using a PC equipped with 
two $2.60$GHz Intel\textsuperscript{\textregistered} 
Xeon\textsuperscript{\textregistered} E$5$-$2640$ v$3$ CPUs.


\subsection{Experimental Analyses}
\label{sec:exp-analyses}
Here, we numerically analyze the behavior of 
our proposed approach.
Due to limited space, we show results on two out of six data sets
and move the rest to Appendix \ref{app:sec:exp-analyses}.


\paragraph{Common Setup:}
As a classifier, we use the Gaussian kernel model:
$g(\bx)=\sum^n_{i=1}w_i \exp(-\|\bx-\bx_i\|^2/(2\sigma^2))$,
where $n=\np+\nn$,
$\{w_i\}^n_{i=1}$ are the parameters, 
$\{\bx_i\}^n_{i=1}=\calXp\cup\calXn$,
and $\sigma>0$ is the Gaussian bandwidth.
The bandwidth candidates are 
$\{1/8, 1/4, 1/2, 1, 3/2, 2\}\times \mathrm{median}
(\|\bx_i-\bx_j\|^{n}_{i,j=1})$.
The classifier trained by minimizing the empirical PN
risk is denoted by  $\widehat{g}_\PN$.
The number of labeled samples for training is $20$, 
where the class-prior was $0.5$.
In all experiments, we used the squared loss for training.
We note that the class-prior of test data was the same as
that of unlabeled data.

\paragraph{Variance Reduction in Practice:}
Here, we numerically investigate 
how many unlabeled samples are sufficient in practice 
such that the variance of the empirical PNU risk is smaller than 
that of the PN risk: 
$\Var[\Rh_\PNU^\eta(g)] < \Var[\Rh_\PN(g)]$ given a fixed classifier $g$. 

As the fixed classifier, we used the classifier $\widehat{g}_\PN$,
where the hyperparameters were determined by five-fold cross-validation.
To compute the variance of the empirical PN and PNU risks,
$\Var[\Rh_\PN(\widehat{g}_\PN)]$ 
and $\Var[\Rh_\PNU^\eta(\widehat{g}_\PN)]$,
we repeatedly drew additional $n_\mathrm{P}^\mathrm{V}=10$ 
positive, $n_\mathrm{N}^\mathrm{V}=10$ negative,
and $n_\mathrm{U}^\mathrm{V}$ unlabeled samples
from the rest of the data set.
The additional samples were also used for
approximating $\widehat{\sigma}_\rP(\widehat{g}_\PN)$
and $\widehat{\sigma}_\rN(\widehat{g}_\PN)$
to compute $\eta$, i.e., 
$\gamma$ in Eqs.\eqref{eq:gam-npnpu} and \eqref{eq:gam-npnnu}.

\begin{figure}[t]
	\centering
	\vspace*{-2mm}
	\subfigure[Phoneme ($d\!=\!5$)]{%
		\includegraphics[clip, width=.5\columnwidth]
		{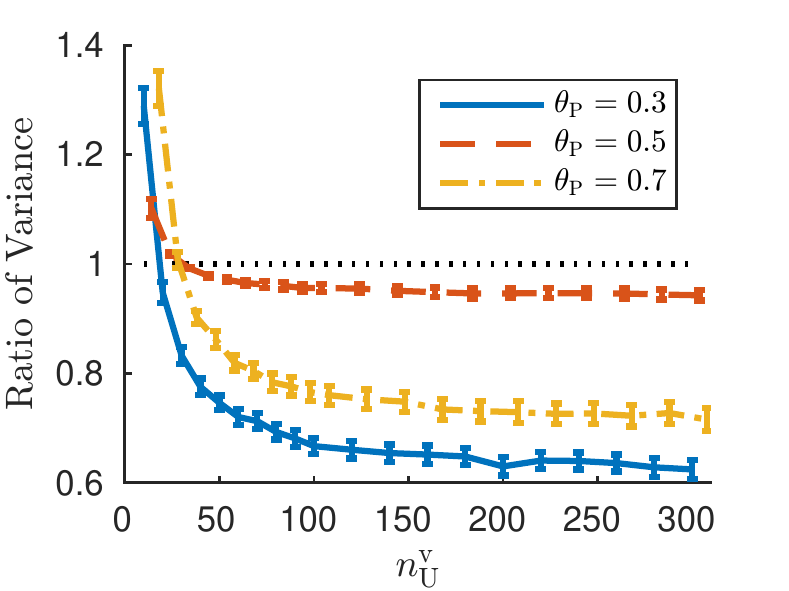}}%
	\subfigure[Magic ($d\!=\!10$)]{%
		\includegraphics[clip, width=.5\columnwidth]
		{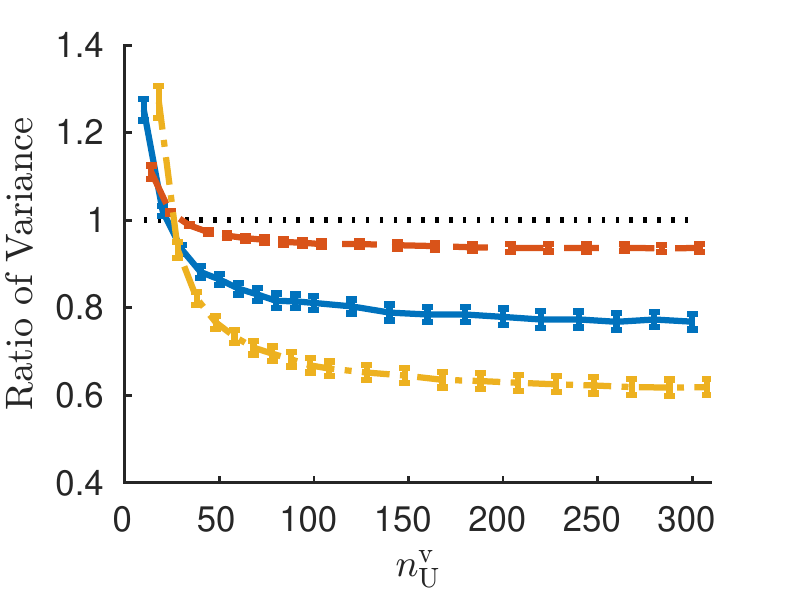}}%
 	\vspace*{-2mm}
	\caption{
	Average and standard error of the ratio 
	between the variance of empirical PNU risk and that of PN risk, 
	$\Var[\Rh_\PNU^\eta(\widehat{g}_\PN)]/\Var[\Rh_\PN(\widehat{g}_\PN)]$,
	as a function of the number of unlabeled samples over $100$ trials.
	Although the variance reduction is proved for an infinite number of samples,
	it can be observed with a finite number of samples.
	}	
 	\vspace*{-1mm}
	\label{fig:var_nu_change}
\end{figure}

Figure \ref{fig:var_nu_change} shows
the ratio between the variance of the empirical PNU risk 
and that of the PN risk, 
$\Var[\Rh_\PNU^\eta(\widehat{g}_\PN)]/\Var[\Rh_\PN(\widehat{g}_\PN)]$.
The number of unlabeled samples for validation 
$n_\mathrm{U}^\mathrm{V}$ increases from $10$ to $300$. 
We see that with a rather small number of unlabeled samples,
the ratio becomes less than $1$.
That is, the variance of the empirical PNU risk becomes smaller than 
that of the PN risk.
This implies that although the variance reduction is proved 
for an infinite number of unlabeled samples,
it can be observed under a finite number of samples in practice.
 
Compared to when $\thetap=0.3$ and $0.7$,  
the effect of variance reduction is small when $\thetap=0.5$.
This is because if we assume $\sigmap(g)\approx\sigman(g)$, 
when $\np\approx\nn$ and $\thetap=0.5$, 
we have $\gamma_\NPNPU\approx\gamma_\NPNNU\approx0$
(because $\psip\approx\psin$. See Theorem~\ref{thm:var-ncvx-pnu}).
That is, the PNU risk is dominated by the PN risk,
implying that $\Var[\Rh_\PNU^\eta(g)]\approx \Var[\Rh_\PN(g)]$.
Note that the class-prior is not the only factor for 
variance reduction; for example, 
if $\thetap=0.5$, $\np\gg\nn$, and $\sigmap(g)\approx\sigman(g)$, 
then $\gamma_\NPNPU\not\approx0$ (because $\psip\ll\psin$) 
and the variance reduction will be large.

\paragraph{PNU Risk in Validation:}
As discussed in Section \ref{sec:theory}, 
the empirical PNU risk will be a reliable validation score
due to its having smaller variance than the empirical PN risk.
We show here that the empirical PNU risk is a promising alternative
to a validation score.

To focus on the effect of validation scores only,
we trained two classifiers by using the same risk, e.g, the empirical PN risk.
We then tune the classifiers with the empirical PN and PNU risks 
denoted by $\widehat{g}_\PN^\PN$ and $\widehat{g}_\PN^\PNU$, respectively. 
The number of validation samples was the same as in the previous experiment.

\begin{figure}[t]
	\centering
	\subfigure[Phoneme ($d\!=\!5$)]{%
		\includegraphics[clip, width=.5\columnwidth]
		{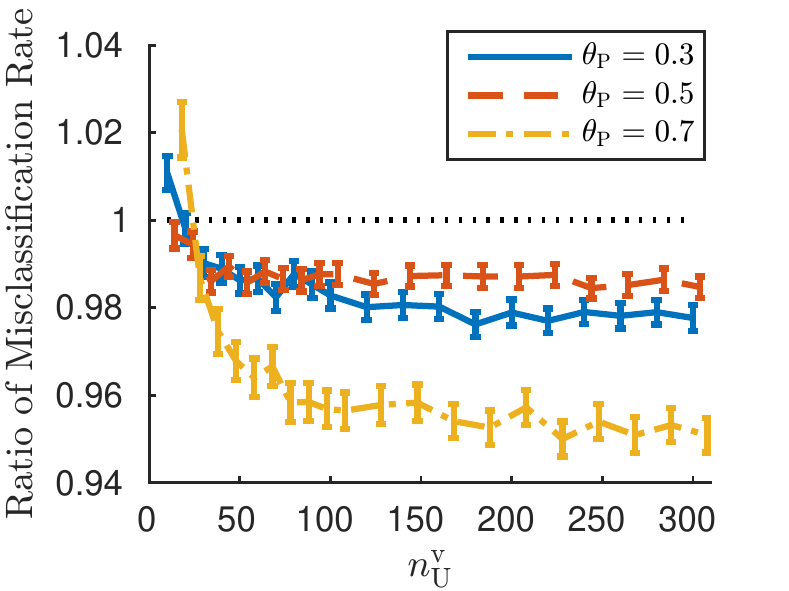}}%
	\subfigure[Magic ($d\!=\!10$)]{%
		\includegraphics[clip, width=.5\columnwidth]
		{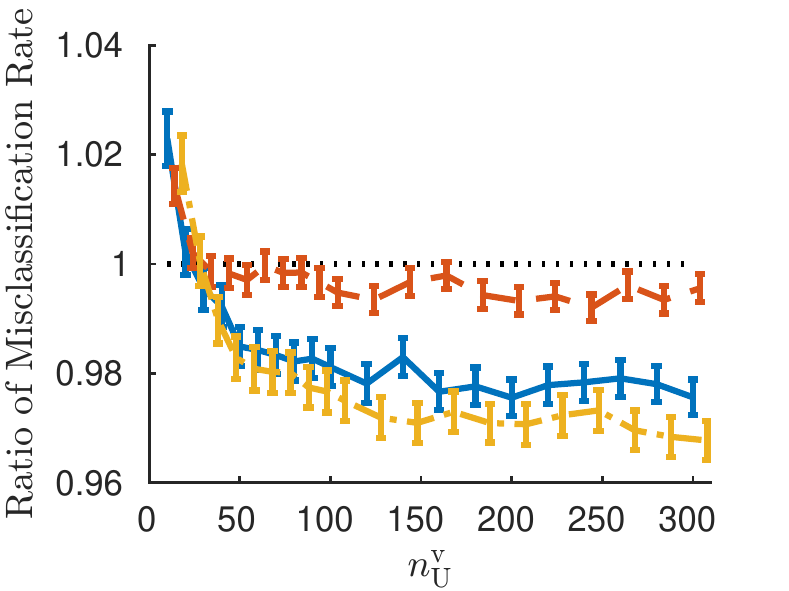}}%
 	\vspace*{-2mm}				
	\caption{
	Average and standard error of the ratio between 
	the misclassification rates of $\widehat{g}^\PNU_\PN$ 
	and $\widehat{g}^\PN_\PN$ 
	as a function of unlabeled samples over $1000$ trials. 
	In many cases, the ratio becomes less than $1$,
	implying that the PNU risk is a promising alternative to 
	the standard PN risk in validation if unlabeled data are available.
	}
 	\vspace*{-1mm}
	\label{fig:val_nu_change}
\end{figure}

Figure~\ref{fig:val_nu_change} 
shows the ratio between the misclassification rate of $\widehat{g}^\PNU_\PN$ 
and that of $\widehat{g}^\PN_\PN$.
The number of unlabeled samples for validation 
increases from $10$ to $300$.
With a rather small number of unlabeled samples,
the ratio becomes less than $1$,
i.e., $\widehat{g}^\PNU_\PN$ achieves better performance than $\widehat{g}^\PN_\PN$.
In particular, when $\thetap=0.3$ and $0.7$, 
$\widehat{g}^\PNU_\PN$ improved substantially;
the large improvement tends to give the large variance reduction 
(cf. Figure \ref{fig:var_nu_change}).
This result shows that the use of the empirical PNU risk for validation 
improved the classification performance given 
a relatively large size of unlabeled data.

\begin{table}[t]
	\centering
	\vspace*{-1mm}
	\caption{Average and standard error of the misclassification rates of each
	method over $50$ trials for benchmark data sets.
    Boldface numbers denote the best and comparable methods in terms of 
	average misclassifications rate
    according to a t-test at a significance level of $5\%$.
    The bottom row gives the number of best/comparable cases of each method.    
	}		
	\vspace*{1mm}
	\label{tab:bench}
	\resizebox{.5\textwidth}{!}{%
 	\begin{tabular}{@{}c@{}r@{\,\,}r@{\,\,}r@{\,\,}r@{\!}r@{\,}r@{\!}r@{\,\,}r}
		\toprule 
		Data set & $\nl$  
 		& \multicolumn{1}{c}{PNU}  & \multicolumn{1}{c}{PUNU}
 		& \multicolumn{1}{c}{ER}    & \multicolumn{1}{c}{LapSVM}
 		& \multicolumn{1}{c}{SMIR} & \multicolumn{1}{c}{WellSVM} 
 		& \multicolumn{1}{c}{S$4$VM} \\
		\toprule
	Banana
	& $10$
& $\mathbf{30.1}$ ($\mathbf{1.0}$) 
& $\mathbf{32.1}$ ($\mathbf{1.1}$) 
& $35.8$ ($1.0$) 
& $36.9$ ($1.0$) 
& $37.7$ ($1.1$) 
& $41.8$ ($0.6$) 
& $45.3$ ($1.0$) 
	\\ 
	$d=2$
	& $50$
& $\mathbf{19.0}$ ($\mathbf{0.6}$) 
& $26.4$ ($1.2$) 
& $20.6$ ($0.7$) 
& $21.3$ ($0.7$) 
& $21.1$ ($1.0$) 
& $42.6$ ($0.5$) 
& $38.7$ ($0.9$) 
	\\ 
	\cmidrule(lr){1-9} 
	Phoneme
	& $10$
& $32.5$ ($0.8$) 
& $33.5$ ($1.0$) 
& $33.4$ ($1.2$) 
& $36.5$ ($1.5$) 
& $36.4$ ($1.2$) 
& $\mathbf{28.4}$ ($\mathbf{0.6}$) 
& $33.7$ ($1.4$) 
	\\ 
	$d=5$
	& $50$
& $28.1$ ($0.5$) 
& $32.8$ ($0.9$) 
& $27.8$ ($0.6$) 
& $27.0$ ($0.8$) 
& $28.6$ ($1.0$) 
& $26.8$ ($0.4$) 
& $\mathbf{25.1}$ ($\mathbf{0.2}$) 
	\\ 
	\cmidrule(lr){1-9} 
	Magic
	& $10$
& $\mathbf{31.7}$ ($\mathbf{0.8}$) 
& $34.1$ ($0.9$) 
& $34.2$ ($1.1$) 
& $37.9$ ($1.3$) 
& $36.0$ ($1.2$) 
& $\mathbf{30.1}$ ($\mathbf{0.8}$) 
& $33.3$ ($0.9$) 
	\\ 
	$d=10$
	& $50$
& $\mathbf{29.9}$ ($\mathbf{0.8}$) 
& $33.4$ ($0.9$) 
& $30.9$ ($0.5$) 
& $31.0$ ($0.9$) 
& $\mathbf{30.8}$ ($\mathbf{0.9}$) 
& $\mathbf{28.8}$ ($\mathbf{0.8}$) 
& $\mathbf{29.2}$ ($\mathbf{0.4}$) 
	\\ 
	\cmidrule(lr){1-9} 
	Image
	& $10$
& $\mathbf{29.8}$ ($\mathbf{0.9}$) 
& $\mathbf{31.7}$ ($\mathbf{0.8}$) 
& $33.7$ ($1.1$) 
& $36.6$ ($1.2$) 
& $36.7$ ($1.2$) 
& $34.7$ ($1.1$) 
& $35.9$ ($1.0$) 
	\\ 
	$d=18$
	& $50$
& $\mathbf{20.7}$ ($\mathbf{0.8}$) 
& $26.6$ ($1.1$) 
& $\mathbf{20.8}$ ($\mathbf{0.8}$) 
& $\mathbf{20.3}$ ($\mathbf{1.0}$) 
& $\mathbf{20.9}$ ($\mathbf{0.9}$) 
& $27.2$ ($1.0$) 
& $23.2$ ($0.7$) 
	\\ 
	\cmidrule(lr){1-9} 
	Susy
	& $10$
& $\mathbf{44.6}$ ($\mathbf{0.6}$) 
& $\mathbf{45.0}$ ($\mathbf{0.6}$) 
& $47.7$ ($0.4$) 
& $48.2$ ($0.4$) 
& $\mathbf{45.1}$ ($\mathbf{0.7}$) 
& $48.0$ ($0.3$) 
& $46.8$ ($0.3$) 
	\\ 
	$d=18$
	& $50$
& $\mathbf{38.9}$ ($\mathbf{0.6}$) 
& $41.5$ ($0.6$) 
& $\mathbf{37.9}$ ($\mathbf{0.7}$) 
& $43.1$ ($0.6$) 
& $43.9$ ($0.8$) 
& $43.8$ ($0.7$) 
& $42.1$ ($0.4$) 
	\\ 
	\cmidrule(lr){1-9} 
	German
	& $10$
& $\mathbf{40.8}$ ($\mathbf{0.9}$) 
& $\mathbf{42.4}$ ($\mathbf{0.7}$) 
& $43.6$ ($0.9$) 
& $45.9$ ($0.7$) 
& $46.2$ ($0.8$) 
& $\mathbf{42.4}$ ($\mathbf{0.8}$) 
& $\mathbf{42.0}$ ($\mathbf{0.7}$) 
	\\ 
	$d=20$
	& $50$
& $\mathbf{36.2}$ ($\mathbf{0.8}$) 
& $39.0$ ($0.8$) 
& $38.9$ ($0.6$) 
& $40.6$ ($0.6$) 
& $38.4$ ($1.1$) 
& $38.5$ ($1.0$) 
& $\mathbf{34.9}$ ($\mathbf{0.5}$) 
	\\ 
	\cmidrule(lr){1-9} 
	Waveform
	& $10$
& $\mathbf{17.4}$ ($\mathbf{0.6}$) 
& $\mathbf{18.0}$ ($\mathbf{0.9}$) 
& $18.5$ ($0.6$) 
& $24.9$ ($1.4$) 
& $\mathbf{18.0}$ ($\mathbf{1.0}$) 
& $\mathbf{16.7}$ ($\mathbf{0.6}$) 
& $20.8$ ($0.8$) 
	\\ 
	$d=21$
	& $50$
& $16.3$ ($0.6$) 
& $23.7$ ($1.2$) 
& $\mathbf{14.2}$ ($\mathbf{0.4}$) 
& $18.1$ ($0.8$) 
& $\mathbf{15.4}$ ($\mathbf{0.6}$) 
& $15.5$ ($0.5$) 
& $15.3$ ($0.3$) 
	\\ 
	\cmidrule(lr){1-9} 
	ijcnn1
	& $10$
& $43.6$ ($0.6$) 
& $\mathbf{40.3}$ ($\mathbf{1.0}$) 
& $49.7$ ($0.1$) 
& $49.2$ ($0.3$) 
& $44.0$ ($1.0$) 
& $45.9$ ($0.7$) 
& $49.3$ ($0.8$) 
	\\ 
	$d=22$
	& $50$
& $\mathbf{34.5}$ ($\mathbf{0.8}$) 
& $37.1$ ($0.9$) 
& $\mathbf{35.5}$ ($\mathbf{0.8}$) 
& $\mathbf{33.4}$ ($\mathbf{1.1}$) 
& $49.4$ ($0.3$) 
& $46.2$ ($0.8$) 
& $48.6$ ($0.4$) 
	\\ 
	\cmidrule(lr){1-9} 
	g50c
	& $10$
& $11.4$ ($0.6$) 
& $12.5$ ($0.6$) 
& $23.3$ ($2.3$) 
& $39.8$ ($1.6$) 
& $21.9$ ($1.3$) 
& $\mathbf{6.6}$ ($\mathbf{0.4}$) 
& $27.0$ ($1.4$) 
	\\ 
	$d=50$
	& $50$
& $12.5$ ($1.1$) 
& $10.1$ ($0.6$) 
& $8.7$ ($0.4$) 
& $22.5$ ($1.5$) 
& $10.6$ ($0.6$) 
& $\mathbf{7.4}$ ($\mathbf{0.4}$) 
& $12.1$ ($0.5$) 
	\\ 
	\cmidrule(lr){1-9} 
	covtype
	& $10$
& $\mathbf{46.2}$ ($\mathbf{0.4}$) 
& $\mathbf{46.0}$ ($\mathbf{0.4}$) 
& $\mathbf{46.0}$ ($\mathbf{0.5}$) 
& $47.1$ ($0.5$) 
& $47.9$ ($0.5$) 
& $\mathbf{46.9}$ ($\mathbf{0.6}$) 
& $\mathbf{46.4}$ ($\mathbf{0.4}$) 
	\\ 
	$d=54$
	& $50$
& $\mathbf{41.3}$ ($\mathbf{0.5}$) 
& $42.3$ ($0.5$) 
& $\mathbf{41.0}$ ($\mathbf{0.4}$) 
& $\mathbf{41.5}$ ($\mathbf{0.5}$) 
& $46.2$ ($0.8$) 
& $43.6$ ($0.6$) 
& $\mathbf{40.8}$ ($\mathbf{0.4}$) 
	\\ 
	\cmidrule(lr){1-9} 
	Spambase
	& $10$
& $27.2$ ($0.9$) 
& $28.1$ ($1.1$) 
& $31.8$ ($1.4$) 
& $39.7$ ($1.4$) 
& $30.9$ ($1.3$) 
& $\mathbf{23.8}$ ($\mathbf{0.8}$) 
& $36.1$ ($1.5$) 
	\\ 
	$d=57$
	& $50$
& $23.4$ ($1.0$) 
& $26.6$ ($1.0$) 
& $22.1$ ($0.7$) 
& $28.5$ ($1.3$) 
& $20.9$ ($0.5$) 
& $\mathbf{19.1}$ ($\mathbf{0.4}$) 
& $24.5$ ($0.9$) 
	\\ 
	\cmidrule(lr){1-9} 
	Splice
	& $10$
& $\mathbf{38.3}$ ($\mathbf{0.8}$) 
& $\mathbf{39.3}$ ($\mathbf{0.8}$) 
& $43.9$ ($0.8$) 
& $47.9$ ($0.5$) 
& $41.6$ ($0.7$) 
& $42.0$ ($1.0$) 
& $42.4$ ($0.6$) 
	\\ 
	$d=60$
	& $50$
& $\mathbf{30.6}$ ($\mathbf{0.8}$) 
& $34.7$ ($0.9$) 
& $\mathbf{30.9}$ ($\mathbf{0.8}$) 
& $38.8$ ($1.0$) 
& $\mathbf{30.6}$ ($\mathbf{0.9}$) 
& $40.9$ ($0.8$) 
& $35.9$ ($0.7$) 
	\\ 
	\cmidrule(lr){1-9} 
	phishing
	& $10$
& $\mathbf{24.2}$ ($\mathbf{1.2}$) 
& $\mathbf{25.8}$ ($\mathbf{1.0}$) 
& $\mathbf{27.3}$ ($\mathbf{1.6}$) 
& $37.2$ ($1.6$) 
& $27.6$ ($1.6$) 
& $27.5$ ($1.4$) 
& $31.7$ ($1.3$) 
	\\ 
	$d=68$
	& $50$
& $\mathbf{15.8}$ ($\mathbf{0.6}$) 
& $18.3$ ($0.8$) 
& $\mathbf{15.4}$ ($\mathbf{0.5}$) 
& $21.1$ ($1.3$) 
& $\mathbf{14.7}$ ($\mathbf{0.8}$) 
& $17.2$ ($0.7$) 
& $16.7$ ($0.8$) 
	\\ 
	\cmidrule(lr){1-9} 
	a9a
	& $10$
& $\mathbf{31.4}$ ($\mathbf{0.9}$) 
& $\mathbf{31.3}$ ($\mathbf{1.0}$) 
& $34.3$ ($1.2$) 
& $41.0$ ($1.1$) 
& $37.3$ ($1.3$) 
& $\mathbf{33.1}$ ($\mathbf{1.2}$) 
& $34.3$ ($1.2$) 
	\\ 
	$d=83$
	& $50$
& $27.9$ ($0.6$) 
& $29.9$ ($0.8$) 
& $28.6$ ($0.7$) 
& $33.3$ ($1.0$) 
& $\mathbf{26.9}$ ($\mathbf{0.7}$) 
& $28.9$ ($0.8$) 
& $\mathbf{26.2}$ ($\mathbf{0.4}$) 
	\\ 
	\cmidrule(lr){1-9} 
	Coil2
	& $10$
& $\mathbf{38.7}$ ($\mathbf{0.8}$) 
& $\mathbf{40.1}$ ($\mathbf{0.8}$) 
& $42.8$ ($0.7$) 
& $43.9$ ($0.8$) 
& $43.2$ ($0.8$) 
& $\mathbf{39.1}$ ($\mathbf{0.9}$) 
& $44.0$ ($0.8$) 
	\\ 
	$d=241$
	& $50$
& $\mathbf{23.2}$ ($\mathbf{0.6}$) 
& $30.5$ ($0.9$) 
& $\mathbf{23.6}$ ($\mathbf{0.9}$) 
& $\mathbf{22.8}$ ($\mathbf{0.9}$) 
& $25.1$ ($0.9$) 
& $\mathbf{22.6}$ ($\mathbf{0.8}$) 
& $25.4$ ($0.8$) 
	\\ 
	\cmidrule(lr){1-9} 
	w8a
	& $10$
& $\mathbf{35.9}$ ($\mathbf{0.9}$) 
& $\mathbf{33.6}$ ($\mathbf{1.0}$) 
& $41.6$ ($1.0$) 
& $46.6$ ($0.8$) 
& $39.4$ ($0.9$) 
& $42.1$ ($0.8$) 
& $43.0$ ($0.8$) 
	\\ 
	$d=300$
	& $50$
& $\mathbf{28.1}$ ($\mathbf{0.7}$) 
& $\mathbf{27.6}$ ($\mathbf{0.6}$) 
& $\mathbf{27.0}$ ($\mathbf{0.9}$) 
& $38.7$ ($0.8$) 
& $\mathbf{28.0}$ ($\mathbf{0.9}$) 
& $33.7$ ($0.8$) 
& $35.2$ ($1.0$) 
	\\ 

 		\cmidrule{1-9}		 	
 		\#Best/Comp.~ &  
 		& $23$ & $13$ & $11$ & $4$ & $9$ & $13$ & $7$ \\
		\bottomrule
	\end{tabular}		
	}
	\vspace*{-3mm}
\end{table}
\begin{figure}[t]
	\centering
	\includegraphics[clip,width=\columnwidth]{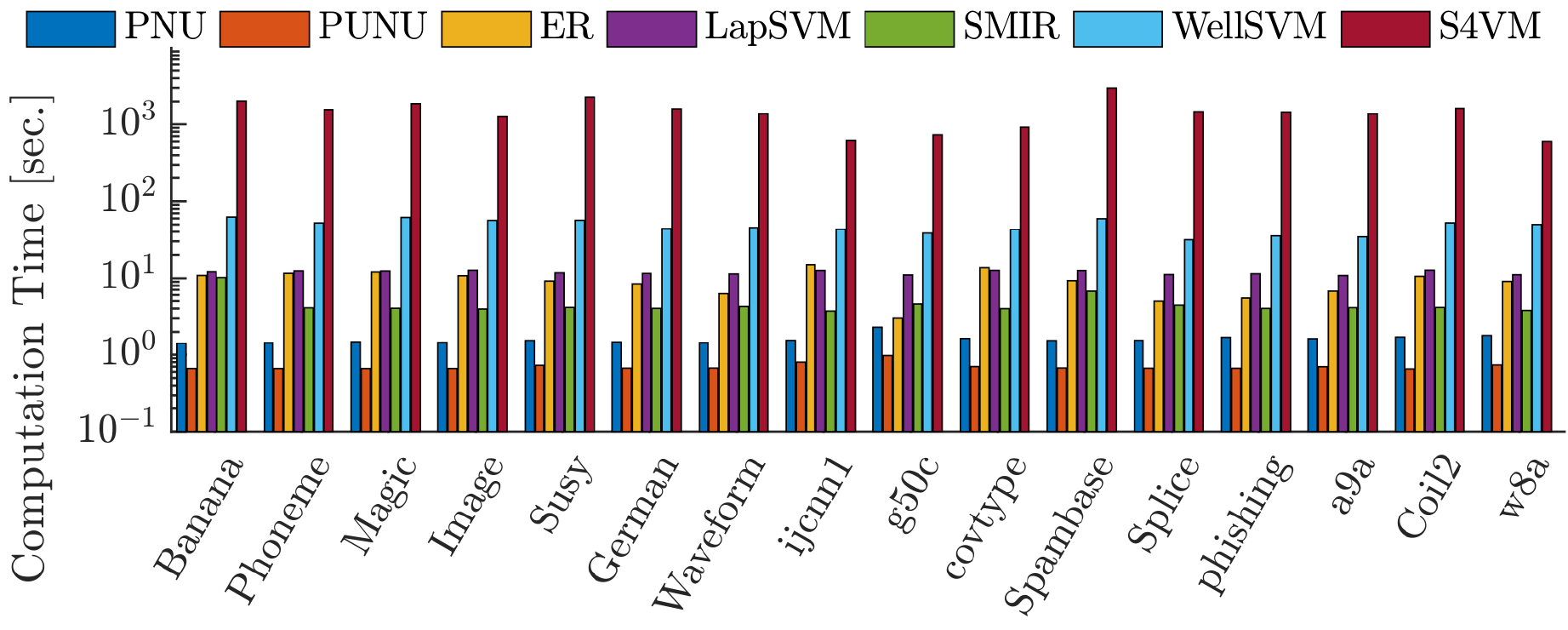}
  	\vspace*{-8mm} 		
	\caption{Average computation time  over $50$ trials  
	for benchmark data sets when $\nl=50$.
	}
	\vspace*{-3mm} 		
	\label{fig:bench_time}
\end{figure}

\subsection{Comparison with Existing Methods}
\label{sec:exp-comp-existing}
Next, we numerically compare the proposed methods
against existing semi-supervised classification methods.

\paragraph{Common Setup:}
We compare our methods against five conventional semi-supervised 
classification methods:
\emph{entropy regularization} (ER) \citep{NIPS:Grandvalet+Bengio:2004},
the \emph{Laplacian support vector machine} (LapSVM) \citep{JMLR:Belkin+etal:2006,JMLR:Melacci+Belkin:2011},
\emph{squared-loss mutual information regularization} (SMIR) \citep{ICML:Gang+etal:2013},
the \emph{weakly labeled support vector machine} (WellSVM) \citep{JMLR:Li+etal:2013},
and the \emph{safe semi-supervised support vector machine} (S$4$VM) \citep{PAMI:Li+Zhou:2015}.

Among the proposed methods,
PNU classification and PUNU classification with the squared loss were
tested.\footnote{
In preliminary experiments,
we tested other loss functions such as the ramp and logistic losses and
concluded that the difference in loss functions did not provide noticeable
difference.
}

\begin{table}[t]		
 	\centering \scriptsize
 	\vspace*{-2mm}
	\caption{Average and standard error of misclassification rates 
	over $30$ trials for the Places $205$ data set.
    Boldface numbers denote the best and comparable methods 
	in terms of the average misclassification rate
    according to a t-test at a significance level of $5\%$.
	}
	\label{tab:places_acc}
 	\resizebox{.5\textwidth}{!}{%
	\begin{tabular}{@{}l@{\,\,}r@{\,\,}r@{\,\,}r@{\,\,}r@{\,\,}r@{\,\,}r@{\,\,}r@{\,\,}r}
		\toprule 
 		Data set 
 		& \multicolumn{1}{c}{$\nun$} 
 		& \multicolumn{1}{c}{$\thetap$}
 		& \multicolumn{1}{c}{$\widehat{\theta}_\mathrm{P}$} 
 		& \multicolumn{1}{c}{PNU} 
 		& \multicolumn{1}{c}{ER} 
 		& \multicolumn{1}{c}{LapSVM} 
 		& \multicolumn{1}{c}{SMIR}
 		& \multicolumn{1}{c}{WellSVM} \\
		\midrule
\multirow{3}{*}{Arts}
& $1000$
& $0.50$ 
& $0.49$ ($0.01$) 
& $\mathbf{27.4}$ ($\mathbf{1.3}$) 
& $\mathbf{26.6}$ ($\mathbf{0.5}$) 
& $\mathbf{26.1}$ ($\mathbf{0.7}$) 
& $40.1$ ($3.9$) 
& $\mathbf{27.5}$ ($\mathbf{0.5}$) 
\\
& $5000$
& $0.50$ 
& $0.50$ ($0.01$) 
& $\mathbf{24.8}$ ($\mathbf{0.6}$) 
& $26.1$ ($0.5$) 
& $26.1$ ($0.4$) 
& $30.1$ ($1.6$) 
& N/A 
\\
& $10000$
& $0.50$ 
& $0.52$ ($0.01$) 
& $\mathbf{25.6}$ ($\mathbf{0.7}$) 
& $\mathbf{25.4}$ ($\mathbf{0.5}$) 
& $\mathbf{25.5}$ ($\mathbf{0.6}$) 
& N/A 
& N/A 
\\
\cmidrule(lr){1-9} 
\multirow{3}{*}{Deserts}
& $1000$
& $0.73$ 
& $0.67$ ($0.01$) 
& $\mathbf{13.0}$ ($\mathbf{0.5}$) 
& $15.3$ ($0.6$) 
& $16.7$ ($0.8$) 
& $17.2$ ($0.8$) 
& $18.2$ ($0.7$) 
\\
& $5000$
& $0.73$ 
& $0.67$ ($0.01$) 
& $\mathbf{13.4}$ ($\mathbf{0.4}$) 
& $\mathbf{13.3}$ ($\mathbf{0.5}$) 
& $16.6$ ($0.6$) 
& $24.4$ ($0.6$) 
& N/A 
\\
& $10000$
& $0.73$ 
& $0.68$ ($0.01$) 
& $\mathbf{13.3}$ ($\mathbf{0.5}$) 
& $\mathbf{13.7}$ ($\mathbf{0.6}$) 
& $16.8$ ($0.8$) 
& N/A 
& N/A 
\\
\cmidrule(lr){1-9}
\multirow{3}{*}{Fields}
& $1000$
& $0.65$ 
& $0.57$ ($0.01$) 
& $\mathbf{22.4}$ ($\mathbf{1.0}$) 
& $26.2$ ($1.0$) 
& $26.6$ ($1.3$) 
& $28.2$ ($1.1$) 
& $26.6$ ($0.8$) 
\\
& $5000$
& $0.65$ 
& $0.57$ ($0.01$) 
& $\mathbf{20.6}$ ($\mathbf{0.5}$) 
& $22.6$ ($0.6$) 
& $24.7$ ($0.8$) 
& $29.6$ ($1.2$) 
& N/A 
\\
& $10000$
& $0.65$ 
& $0.57$ ($0.01$) 
& $\mathbf{21.6}$ ($\mathbf{0.6}$) 
& $\mathbf{22.5}$ ($\mathbf{0.6}$) 
& $25.0$ ($0.9$) 
& N/A 
& N/A 
\\
\cmidrule(lr){1-9} 
\multirow{3}{*}{Stadiums}
& $1000$
& $0.50$ 
& $0.50$ ($0.01$) 
& $\mathbf{11.4}$ ($\mathbf{0.4}$) 
& $\mathbf{11.5}$ ($\mathbf{0.5}$) 
& $12.5$ ($0.5$) 
& $\mathbf{17.4}$ ($\mathbf{3.6}$) 
& $\mathbf{11.7}$ ($\mathbf{0.4}$) 
\\
& $5000$
& $0.50$ 
& $0.50$ ($0.01$) 
& $\mathbf{11.0}$ ($\mathbf{0.5}$) 
& $\mathbf{10.9}$ ($\mathbf{0.3}$) 
& $\mathbf{11.1}$ ($\mathbf{0.3}$) 
& $13.4$ ($0.7$) 
& N/A 
\\
& $10000$
& $0.50$ 
& $0.51$ ($0.00$) 
& $\mathbf{10.7}$ ($\mathbf{0.3}$) 
& $\mathbf{10.9}$ ($\mathbf{0.3}$) 
& $\mathbf{11.2}$ ($\mathbf{0.2}$) 
& N/A 
& N/A 
\\
\cmidrule(lr){1-9}
\multirow{3}{*}{Platforms}
& $1000$
& $0.27$ 
& $0.33$ ($0.01$) 
& $\mathbf{21.8}$ ($\mathbf{0.5}$) 
& $23.9$ ($0.6$) 
& $24.1$ ($0.5$) 
& $30.1$ ($2.3$) 
& $26.2$ ($0.8$) 
\\
& $5000$
& $0.27$ 
& $0.34$ ($0.01$) 
& $\mathbf{23.3}$ ($\mathbf{0.8}$) 
& $\mathbf{24.4}$ ($\mathbf{0.7}$) 
& $\mathbf{24.9}$ ($\mathbf{0.7}$) 
& $26.6$ ($0.3$) 
& N/A 
\\
& $10000$
& $0.27$ 
& $0.34$ ($0.01$) 
& $\mathbf{21.4}$ ($\mathbf{0.5}$) 
& $24.3$ ($0.6$) 
& $24.8$ ($0.5$) 
& N/A 
& N/A 
\\
\cmidrule(lr){1-9} 
\multirow{3}{*}{Temples}
& $1000$
& $0.55$ 
& $0.51$ ($0.01$) 
& $\mathbf{43.9}$ ($\mathbf{0.7}$) 
& $\mathbf{43.9}$ ($\mathbf{0.6}$) 
& $\mathbf{43.4}$ ($\mathbf{0.6}$) 
& $50.7$ ($1.6$) 
& $\mathbf{44.3}$ ($\mathbf{0.5}$) 
\\
& $5000$
& $0.55$ 
& $0.54$ ($0.01$) 
& $\mathbf{43.4}$ ($\mathbf{0.9}$) 
& $\mathbf{43.0}$ ($\mathbf{0.6}$) 
& $\mathbf{43.1}$ ($\mathbf{1.0}$) 
& $\mathbf{43.6}$ ($\mathbf{0.7}$) 
& N/A 
\\
& $10000$
& $0.55$ 
& $0.50$ ($0.01$) 
& $\mathbf{45.2}$ ($\mathbf{0.8}$) 
& $\mathbf{44.4}$ ($\mathbf{0.8}$) 
& $\mathbf{44.2}$ ($\mathbf{0.7}$) 
& N/A 
& N/A 
\\
		\bottomrule
	\end{tabular}
 	}
 	\vspace*{-4mm}
\end{table}

\paragraph{Data Sets:}
We used sixteen benchmark data sets
taken from the \emph{UCI Machine Learning Repository} \citep{UCI:Lichman:2013},
the \emph{Semi-Supervised Learning book} \citep{book:Chapelle+etal:2006},
the \emph{LIBSVM} \citep{LibSVM:Chang+etal:2011},
the \emph{ELENA Project},\footnote{
\url{https://www.elen.ucl.ac.be/neural-nets/Research/Projects/ELENA/elena.htm}}
and a paper by \citet{AISTATS:Chapelle+Zien:2005}.\footnote{
\url{http://olivier.chapelle.cc/lds/}}
Each feature was scaled to $[0, 1]$.
Similarly to the setting in Section~\ref{sec:exp-analyses},
we used the Gaussian kernel model for all methods.
The training data is $\{\bx_i\}^n_{i=1}=\calXp\cup\calXn\cup\calXu$,
where $n=\np+\nn+\nun$.
We selected all hyper-parameters with validation samples of size
$20$ ($n_\mathrm{P}^\mathrm{V}=n_\mathrm{N}^\mathrm{V}=10$).
For training, we drew $\nl$ labeled and $\nun=300$ unlabeled samples.
The class-prior of labeled data was set at $0.7$ 
and that of unlabeled samples was set at $\thetap=0.5$ 
that were assumed to be known. In practice, 
the class-prior, $\thetap$, can be estimated by methods proposed, e.g., by 
\citet{JMLR:Blanchard+etal:2010},
\citet{ICML:Ramaswamy+etal:2016},
or \citet{IEICE:Kawakubo+etal:2016}.

Table \ref{tab:bench} lists the average and standard error of 
the misclassification rates over $50$ trials
and the number of best/comparable performances of each method in the bottom row.
The superior performance of PNU classification over PUNU classification
agrees well with the discussion in Section~\ref{sec:theory-punu-pnu}.
With the g$50$c data set, which well satisfies
the low-density separation principle, the WellSVM 
achieved the best performance.
However, in the Banana data set, where the two classes are highly
overlapped, the performance of WellSVM was worse than 
the other methods.
In contrast, PNU classification achieved consistently
better/comparable performance and its performance 
did not degenerate considerably across data sets.
These results show that the idea of using PU classification
in semi-supervised classification is promising.

Figure~\ref{fig:bench_time} plots the computation time,
which shows that the fastest computation was achieved using 
the proposed methods with the square loss.

\paragraph{Image Classification:}
Finally, we used the 
\emph{Places $205$ data set} \citep{NIPS:Zhou+etal:2014},
which contains $2.5$ million images in $205$ scene classes. 
We used a $4096$-dimensional feature vector extracted from each image
by \emph{AlexNet} under the framework of \emph{Caffe},\footnote{
\url{http://caffe.berkeleyvision.org/}} 
which is available on the project
website\footnote{\url{http://places.csail.mit.edu/}}.
We chose two similar scenes to construct binary classification tasks 
(see the description of data sets in Appendix \ref{sec:desc-places}).
We drew $100$ labeled and $\nun$ unlabeled samples from each task;
the class-prior of labeled and unlabeled data were respectively set at $0.5$ 
and $\thetap=m_\mathrm{P}/(m_\mathrm{P} + m_\mathrm{N})$,
where $m_\mathrm{P}$ and $m_\mathrm{N}$ 
respectively denote the number of total samples
in positive and negative scenes.
We used a linear classifier $g(\bx)=\bw^\top\bx + w_0$,
where $\bw$ is the weight vector and $w_0$ is the offset
(in the SMIR, the linear kernel model is used; 
see \citet{ICML:Gang+etal:2013} for details).

We selected hyper-parameters in PNU classification
by applying five-fold cross-validation with respect to
$R^{\bar{\eta}}_\PNU(g)$ with the zero-one loss,
where
$\bar{\eta}$ was set at Eq.\eqref{eq:gam-npnpu} or Eq.\eqref{eq:gam-npnnu}
with $\sigmap(g)=\sigman(g)$.
The class-prior $p(y=+1)=\thetap$ was estimated using the method based on
energy distance minimization \citep{IEICE:Kawakubo+etal:2016}.

\begin{figure}[t]
 	\centering
 	\includegraphics[clip,width=\columnwidth]{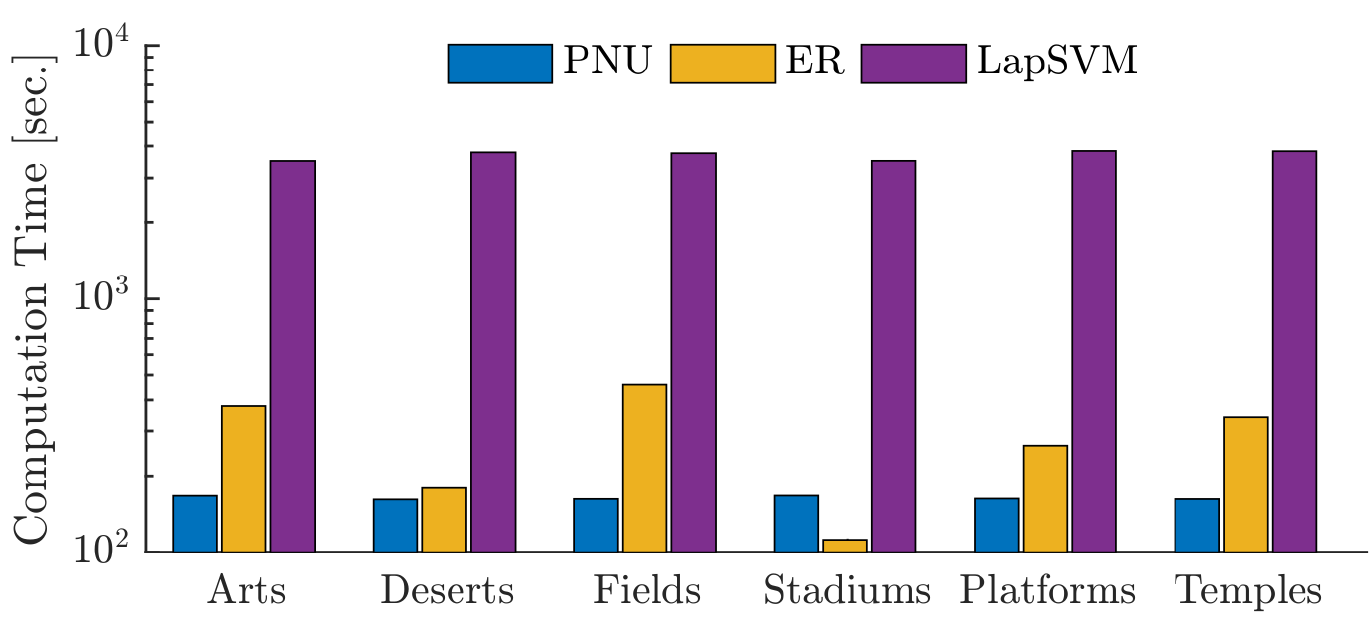}
	\vspace*{-7mm} 	
 	\caption{Average computation time over $30$ trials
           for the Places $205$ data set when $n_\un=10000$.
           }
	\vspace*{-3mm}           
 	\label{fig:places_time}  	
\end{figure}

Table~\ref{tab:places_acc} lists the average and standard error of the
misclassification rates over $30$ trials, 
where methods taking more than $2$ hours were omitted 
and indicated as \emph{N/A}.
The results show that PNU classification was most effective.
The average computation times are shown in  
Figure~\ref{fig:places_time},
revealing again that PNU classification was the fastest method.

\section{Conclusions}
\label{sec:conclusions}
In this paper, we proposed a novel semi-supervised classification
approach based on classification from positive and unlabeled data.
Unlike most of the conventional methods, our approach does not require 
strong assumptions on the data distribution
such as the cluster assumption.
We theoretically analyzed the variance of risk estimators
and showed that unlabeled data help reduce the variance
without the conventional distributional assumptions.
We also established generalization error bounds
and showed that the confidence term decreases 
with respect to the number of positive, negative, and unlabeled samples
without the conventional distributional assumptions
in the optimal parametric order. 
We experimentally analyzed the behavior of the proposed methods
and demonstrated that one of the proposed methods, 
termed PNU classification, was most effective
in terms of both classification accuracy and computational efficiency.
It was recently pointed out that PU classification can behave
undesirably for very flexible models and a modified PU risk has been
proposed \citep{arXiv:Kiryo+etal:2017}.
Our future work is to develop a semi-supervised
classification method based on the modified PU classification.



\newpage
\section*{Acknowledgements}
TS was supported by JSPS KAKENHI $15$J$09111$.
GN was supported by the JST CREST program and Microsoft Research Asia.
MCdP and MS were supported by the JST CREST program.

\bibliographystyle{icml2017}
\bibliography{pnu_short}

\onecolumn
\appendix

\section{Proofs of Theorems}
\label{sec:proofs}
In this section, we give the proofs of Theorems in Section~\ref{sec:theory}.

\subsection{Proof of Theorem~\ref{thm:gen-err-rl}}

Recall that
\begin{align*}
R_\NPUNU^\gamma(g)
&= (1-\gamma) R_\NPU(g) + \gamma R_\NNU(g)\\
&= (2-2\gamma)\thetap\Rp(g) +2\gamma\thetan\Rn(g) 
+(1-\gamma)\Run(g) +\gamma\Rup(g) +\const,\\
R_\NPNPU^\gamma(g)
&= (1-\gamma) R_\PN(g) + \gamma R_\NPU(g)\\
&= (1+\gamma)\thetap\Rp(g) +(1-\gamma)\thetan\Rn(g) 
+\gamma\Run(g) +\const,\\
R_\NPNNU^\gamma(g)
&= (1-\gamma) R_\PN(g) + \gamma R_\NNU(g)\\
&= (1-\gamma)\thetap\Rp(g) +(1+\gamma)\thetan\Rn(g) 
+\gamma\Rup(g) +\const.
\end{align*}
Let $\hRp(g)$, $\hRn(g)$, $\hRup(g)$ and $\hRun(g)$ be the empirical risks. 
In order to prove Theorem~\ref{thm:gen-err-rl}, the following concentration lemma is needed:

\begin{lemma}
  \label{thm:uni-dev-marginal-risks-rl}%
  For any $\delta>0$, we have these uniform deviation bounds with probability at least $1-\delta/3$:
  \begin{align*}
  \sup\nolimits_{g\in\calG}(\Rp(g)-\hRp(g))
  &\le \frac{C_wC_\phi}{\sqrt{\np}} +\sqrt{\frac{\ln(3/\delta)}{2\np}},\\
  \sup\nolimits_{g\in\calG}(\Rn(g)-\hRn(g))
  &\le \frac{C_wC_\phi}{\sqrt{\nn}} +\sqrt{\frac{\ln(3/\delta)}{2\nn}},\\
  \sup\nolimits_{g\in\calG}(\Rup(g)-\hRup(g))
  &\le \frac{C_wC_\phi}{\sqrt{\nun}} +\sqrt{\frac{\ln(3/\delta)}{2\nun}},\\
  \sup\nolimits_{g\in\calG}(\Run(g)-\hRun(g))
  &\le \frac{C_wC_\phi}{\sqrt{\nun}} +\sqrt{\frac{\ln(3/\delta)}{2\nun}}.
  \end{align*}
\end{lemma}

All inequalities in Lemma~\ref{thm:uni-dev-marginal-risks-rl} are from 
the basic \emph{uniform deviation bound} using the Rademacher complexity \citep{book:Mohri+etal:2012},
\emph{Talagrand's contraction lemma} \citep{book:Ledoux+Talagrand:1991}, 
as well as the fact that the Lipschitz constant of $\ellR$ is $1/2$. 
For these reasons, the detailed proof of Lemma~\ref{thm:uni-dev-marginal-risks-rl} is omitted.

Consider $R_\NPNPU^\gamma(g)$. It is clear that
\begin{align*}
&\sup\nolimits_{g\in\calG}(R_\NPNPU^\gamma(g)-\Rh_\NPNPU^\gamma(g)) \\ 
&\le (1+\gamma)\thetap\sup\nolimits_{g\in\calG}(\Rp(g)-\hRp(g))
+(1-\gamma)\thetan\sup\nolimits_{g\in\calG}(\Rn(g)-\hRn(g)) 
+\gamma\sup\nolimits_{g\in\calG}(\Run(g)-\hRun(g)).
\end{align*}
Therefore, by applying Lemma~\ref{thm:uni-dev-marginal-risks-rl}, for any $\delta > 0$, 
it holds with probability at least $1-\delta$ that
\begin{align*}
\sup\nolimits_{g\in\calG}&(R_\NPNPU^\gamma(g)-\Rh_\NPNPU^\gamma(g)) 
\le \frac{1}{2}C_{w,\phi,\delta}\cdot\chi(1+\gamma,1-\gamma,\gamma).
\end{align*}
Since $I(g)\le2R_\NPNPU^\gamma$, with the same probability,
\begin{align*}
\sup\nolimits_{g\in\calG}(I(g)\!-\!2\Rh_\NPNPU^\gamma(g))
\le C_{w,\phi,\delta}\!\cdot\!\chi(1+\gamma,1-\gamma,\gamma).
\end{align*}
Similarly, $\sup_{g\in\calG}(I(g)-2\Rh_\NPNNU^\gamma(g))\le C_{w,\phi,\delta}\cdot\chi(1-\gamma,1+\gamma,\gamma)$ with probability 
at least $1-\delta$.

Finally, $R_\NPUNU^\gamma(g)$ is slightly more involved, for that there are both $\Rup(g)$ and $\Run(g)$. 
From $\ellR(m)+\ellR(-m)=1$, we can know $\Rup(g)+\Run(g)=1$ and then
\begin{align*}
(1-\gamma)\Run(g)+\gamma\Rup(g) 
&=
\begin{cases}
(2\gamma-1)\Rup(g)+\const & \gamma\ge1/2, \\
(1-2\gamma)\Run(g)+\const & \gamma<1/2.
\end{cases}
\end{align*}
As a result, $\sup_{g\in\calG}(I(g)-2\Rh_\NPUNU^\gamma(g))\le C_{w,\phi,\delta}\cdot\chi(2-2\gamma,2\gamma,|2\gamma-1|)$ 
with probability at least $1-\delta$.

\subsection{Proof of Theorem~\ref{thm:gen-err-sl}}

In fact,
\begin{align*}
\ellTS(m)=
\begin{cases}
1/4 & m\le0,\\
(m-1)^2/4 & 0<m\le1,\\
0 & m>1,
\end{cases}
\end{align*}
and after plugging this $\ellTS(m)$ into $\ellt_\mathrm{TS}(m)$,
\begin{align*}
\ellt_\mathrm{TS}(m)
&=\ellTS(m)-\ellTS(-m)\\
&=\begin{cases}
1/4 & m\le-1,\\
1/4-(m+1)^2/4 & -1<m\le0,\\
(m-1)^2/4-1/4 & 0<m\le1,\\
-1/4 & m>1.
\end{cases}
\end{align*}
It is easy to see that $\ellTS(m)$ and 
$\ellt_\mathrm{TS}(m)$ are Lipschitz
continuous with the same Lipschitz constant $1/2$.

Next, recall that
\begin{align*}
R_\CPUNU^\gamma(g)
&= (1-\gamma) R_\CPU(g) + \gamma R_\CNU(g)\\
&= (1-\gamma)\thetap\Rp'(g) +\gamma\thetan\Rn'(g) 
+(1-\gamma)\Run(g) +\gamma\Rup(g),\\
R_\CPNPU^\gamma(g)
&= (1-\gamma) R_\PN(g) + \gamma R_\CPU(g)\\
&= (1-\gamma)\thetap\Rp(g) +(1-\gamma)\thetan\Rn(g) 
+\gamma\thetap\Rp'(g) +\gamma\Run(g),\\
R_\CPNNU^\gamma(g)
&= (1-\gamma) R_\PN(g) + \gamma R_\CNU(g)\\
&= (1-\gamma)\thetap\Rp(g) +(1-\gamma)\thetan\Rn(g)
+\gamma\thetan\Rn'(g) +\gamma\Rup(g).
\end{align*}
Let $\hRp(g)$, $\hRn(g)$, $\hRup(g)$, $\hRun(g)$, $\hRp'(g)$ and $\hRn'(g)$ be the empirical risks. Again, the following concentration lemma is needed:

\begin{lemma}
  \label{thm:uni-dev-marginal-risks-tsl}%
  For any $\delta>0$, we have these uniform deviation bounds with probability at least $1-\delta/4$:
  \begin{align*}
  \sup\nolimits_{g\in\calG}(\Rp(g)-\hRp(g))
  &\le \frac{C_wC_\phi}{\sqrt{\np}} +\sqrt{\frac{\ln(4/\delta)}{32\np}},\\
  \sup\nolimits_{g\in\calG}(\Rn(g)-\hRn(g))
  &\le \frac{C_wC_\phi}{\sqrt{\nn}} +\sqrt{\frac{\ln(4/\delta)}{32\nn}},\\
  \sup\nolimits_{g\in\calG}(\Rup(g)-\hRup(g))
  &\le \frac{C_wC_\phi}{\sqrt{\nun}} +\sqrt{\frac{\ln(4/\delta)}{32\nun}},\\
  \sup\nolimits_{g\in\calG}(\Run(g)-\hRun(g))
  &\le \frac{C_wC_\phi}{\sqrt{\nun}} +\sqrt{\frac{\ln(4/\delta)}{32\nun}},\\
  \sup\nolimits_{g\in\calG}(\Rp'(g)-\hRp'(g))
  &\le \frac{C_wC_\phi}{\sqrt{\np}} +\sqrt{\frac{\ln(4/\delta)}{8\np}},\\
  \sup\nolimits_{g\in\calG}(\Rn'(g)-\hRn'(g))
  &\le \frac{C_wC_\phi}{\sqrt{\nn}} +\sqrt{\frac{\ln(4/\delta)}{8\nn}}.
  \end{align*}
\end{lemma}

The detailed proof of Lemma~\ref{thm:uni-dev-marginal-risks-tsl} is omitted for 
the same reason as Lemma~\ref{thm:uni-dev-marginal-risks-rl}. 
The difference is due to that $0\le\ellTS(m)\le1/4$ and 
$-1/4\le\ellt_\mathrm{TS}(m)\le1/4$ 
whereas $0\le\ellR(m)\le1$ just like $0\le\ellzo(m)\le1$. 
For convenience, 
we will relax $1/32$ to $1/8$ in the square root for $\Rp(g)$, $\Rn(g)$, $\Rup(g)$, $\Run(g)$.

Consider $R_\CPUNU^\gamma(g)$. By applying Lemma~\ref{thm:uni-dev-marginal-risks-tsl}, 
for any $\delta > 0$, it holds with probability at least $1-\delta$ that
\begin{align*}
\sup\nolimits_{g\in\calG}&(R_\CPUNU^\gamma(g)-\Rh_\CPUNU^\gamma(g)) 
\le \frac{1}{4}C'_{w,\phi,\delta}\cdot\chi(1-\gamma,\gamma,1).
\end{align*}
Since $I(g)\le4R_\CPUNU^\gamma$, with the same probability,
\begin{align*}
\sup\nolimits_{g\in\calG}(I(g)-4\Rh_\CPUNU^\gamma(g))
\le C'_{w,\phi,\delta}\cdot\chi(1-\gamma,\gamma,1).
\end{align*}
The other two generalization error bounds can be proven similarly.

\subsection{Proofs of Theorems~\ref{thm:var-ncvx-punu} and \ref{thm:var-ncvx-pnu}}

Note that $g$ is independent of the data for evaluating $\Rh_\NPUNU^\gamma(g)$, 
since it is fixed in the evaluation. 
Thus, $\Varp[\hRp(g)]=\sigmap^2(g)/\np$ 
and $\Varn[\hRn(g)]=\sigman^2(g)/\nn$. 
When $\nun\rightarrow\infty$,
\begin{align*}
\Var[\Rh_\NPUNU^\gamma(g)] 
&=4(1-\gamma)^2\thetap^2\Varp[\hRp(g)]
+4\gamma^2\thetan^2\Varn[\hRn(g)]\\
&=4(1-\gamma)^2\psip+4\gamma^2\psin\\
&=4(\psip+\psin)\gamma^2-8\psip\gamma+4\psip,
\end{align*}
and it is obvious that $\gamma_\NPUNU\in[0,1]$. 
All other claims in Theorem~\ref{thm:var-ncvx-punu} follow from 
that $\Var[\Rh_\NPUNU^\gamma(g)]$ is quadratic in $\gamma$, 
that $\Var[\Rh_\NPUNU^\gamma(g)]=\Var[\Rh_\PN(g)]$ at $\gamma=1/2$, 
and that $\gamma_\NPUNU<1/2$ if $\psip<\psin$ or $\gamma_\NPUNU>1/2$ if $\psip>\psin$.

Likewise, when $\nun\rightarrow\infty$,
\begin{align*}
\Var[\Rh_\NPNPU^\gamma(g)]
&=(1+\gamma)^2\psip+(1-\gamma)^2\psin,\\
\Var[\Rh_\NPNNU^\gamma(g)]
&=(1-\gamma)^2\psip+(1+\gamma)^2\psin,
\end{align*}
and $\gamma_\NPNPU\ge0$ if $\psip\le\psin$ or $\gamma_\NPNNU\ge0$ if $\psip\ge\psin$. 
The rest of proof of Theorem~\ref{thm:var-ncvx-pnu} is analogous to that of 
Theorem~\ref{thm:var-ncvx-punu}.

\newpage
\section{Experimental Setting}
Here, we summarized the experimental settings.

\subsection{Implementation in Our Experiments}
We implemented the ER by ourselves, and for the other methods, we used
the codes available at the authors' websites:
\begin{itemize}
\item LapSVM: \url{http://www.dii.unisi.it/~melacci/lapsvmp/}

\item SMIR: \url{http://www.ms.k.u-tokyo.ac.jp/software/SMIR.zip}

\item WellSVM: \url{http://lamda.nju.edu.cn/code_WellSVM.ashx}

\item S$4$VM: \url{http://lamda.nju.edu.cn/files/s4vm.rar}.
\end{itemize}
Note that we modified the original code of the S$4$VM for transductive learning
to inductive learning according to \citet{PAMI:Li+Zhou:2015}. 


\subsection{Parameter Candidates in Our Experiments}
The regularization parameters for all the methods
were chosen from $\{10^{-5},10^{-4},\ldots,10^2\}$,
except the regularization parameter of the SMIR for the squared loss mutual
information (SMI)
and that of the S$4$VM for labeled data.
The number of nearest-neighbors to construct Laplacian matrix
for the LapSVM was chosen from the candidates $\{5, 6, \ldots, 10\}$.
The combination parameter $\eta$ of PNU classification was chosen 
from $\{-1, -0.9, \ldots, 1\}$, 
and $\gamma$ of PUNU classification was chosen from $\{0,0.05,\ldots,1\}$.
We chose these hyper-parameters by five-fold cross-validation.
The parameter for the $\ell_2$-regularizer of the SMIR is set at
$\gamma_\mathrm{S}/(n\cdot \min_{k\in\{\pm 1\}} p(y=k))+0.001$,
where $\gamma_S$ is the regularization parameter for the SMI.
The regularization parameter of the S$4$VM for the labeled data
is set at $1$.
The other parameters were set at the default values.


\subsection{Data Set Description of Image Classification Data Set}
\label{sec:desc-places}
Table \ref{tab:desc_places} is the description of the data sets 
used in the image classification experiment.

\begin{table}[h]		
 	\centering \small
	\caption{
	The description of the data set used in the image classification
	experiment. 
	}
	\label{tab:desc_places}
	\begin{tabular}{ll@{}r}
		\toprule 
 		Data set & Data sources & \multicolumn{1}{l}{\#Samples} \\
		\midrule
\multirow{3}{*}{Arts}
& Art Gallery & ($m_\mathrm{P}\!=\!15000$)
\\
& \;vs. 
\\
& Art Studio & ($m_\mathrm{N}\!=\!15000$) 
\\
\cmidrule(lr){1-3} 
\multirow{3}{*}{Deserts}
& Desert Sand & $(m_\mathrm{P}\!=\!15000)$ 
\\
& \;vs. 
\\
& Desert Vegetation & $(m_\mathrm{N}\!=\!5556)$ 
\\
\cmidrule(lr){1-3}
\multirow{3}{*}{Fields}
& Field Wild & $(m_\mathrm{P}=15000)$ 
\\
& \;vs. 
\\
& Field Cultivated & $(m_\mathrm{N}=8117)$ 
\\
\cmidrule(lr){1-3} 
\multirow{3}{*}{Stadiums}
& Stadium Baseball & $(m_\mathrm{P}\!=\!15000)$ 
\\
& \;vs. 
\\
& Stadium Football & $(m_\mathrm{N}\!=\!15000)$ 
\\
\cmidrule(lr){1-3}
\multirow{3}{*}{Platforms}
& Subway Station & $(m_\mathrm{P}=5597)$ 
\\
& \;vs. 
\\
& Train Station & $(m_\mathrm{N}\!=\!15000)$ 
\\
\cmidrule(lr){1-3} 
\multirow{3}{*}{Temples}
& Temple East Asia & $(m_\mathrm{P}\!=\!8691)$ 
\\
& \;vs. 
\\
& Temple South Asia & $(m_\mathrm{N}\!=\!7178)$ 
\\
		\bottomrule
	\end{tabular}
\end{table}

\newpage
\section{Supplementary Results for Experimental Analyses}
\label{app:sec:exp-analyses}
\begin{figure}[h]
	\centering
	\subfigure[Banana ($d\!=\!2$)]{%
		\includegraphics[clip, width=.33\columnwidth]
		{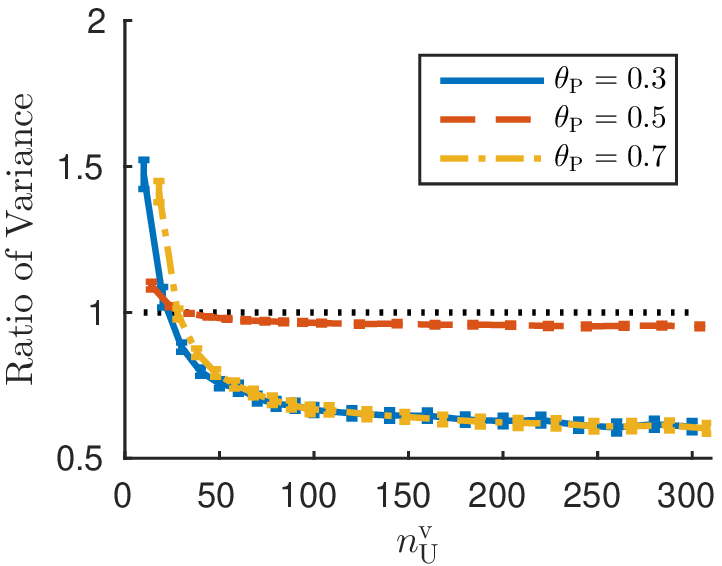}}%
	\subfigure[Waveform ($d\!=\!21$)]{%
		\includegraphics[clip, width=.33\columnwidth]
		{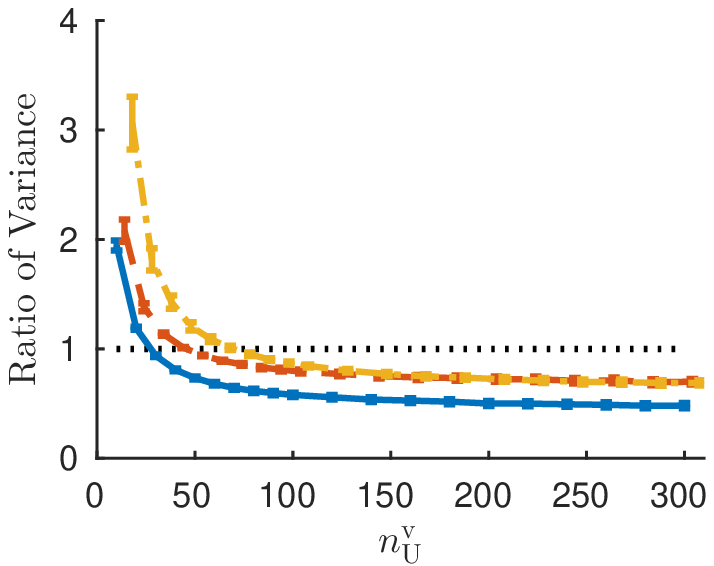}}%
	\subfigure[Splice ($d\!=\!60$)]{%
 		\includegraphics[clip, width=.33\columnwidth]
 		{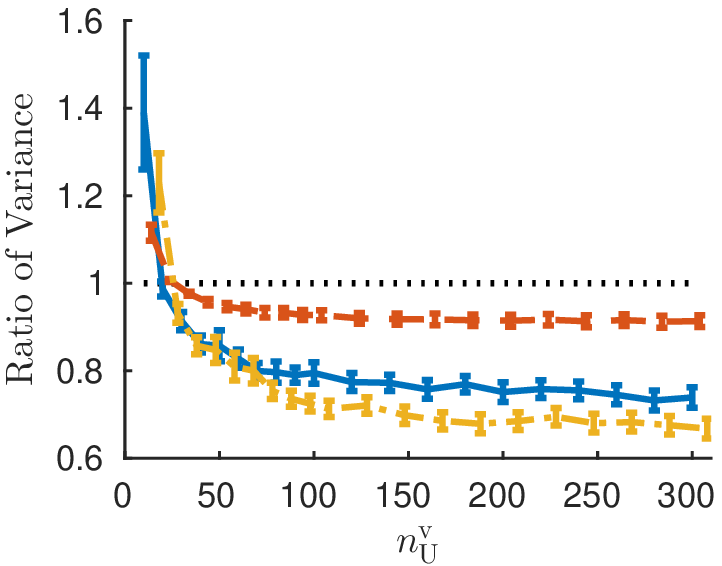}}%
	\caption{
	Average and standard error of the ratio 
	between the variance of the empirical PNU risk and that of the PN risk, 
	$\Var[\Rh_\PNU^\eta(\widehat{g}_\PN)]/\Var[\Rh_\PN(\widehat{g}_\PN)]$,
	as a function of the number of unlabeled samples over $100$ trials.
	Although the variance reduction is proved for an infinite number of samples,
	it can be observed with a finite number of samples.
	}	
	\label{app:fig:var_nu_change}
\end{figure}

\begin{figure}[h]
	\centering
	\subfigure[Banana ($d\!=\!2$)]{%
		\includegraphics[clip, width=.33\columnwidth]
		{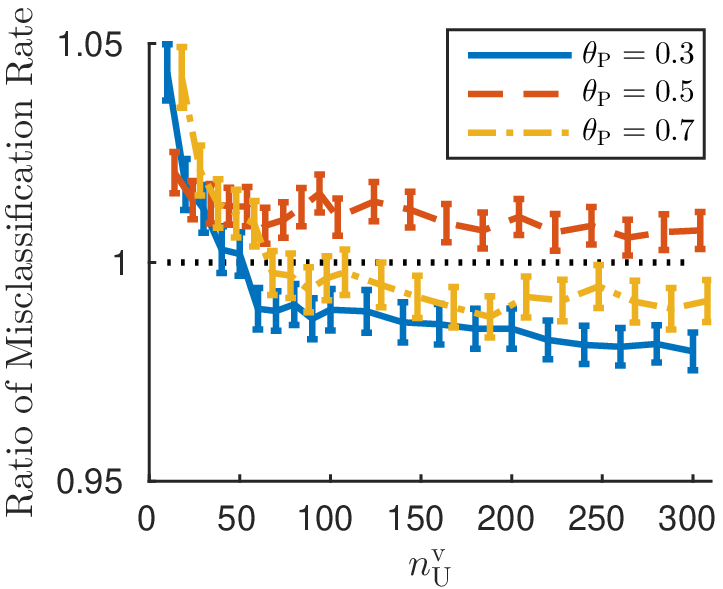}}%
	\subfigure[Waveform ($d\!=\!21$)]{%
		\includegraphics[clip, width=.33\columnwidth]
		{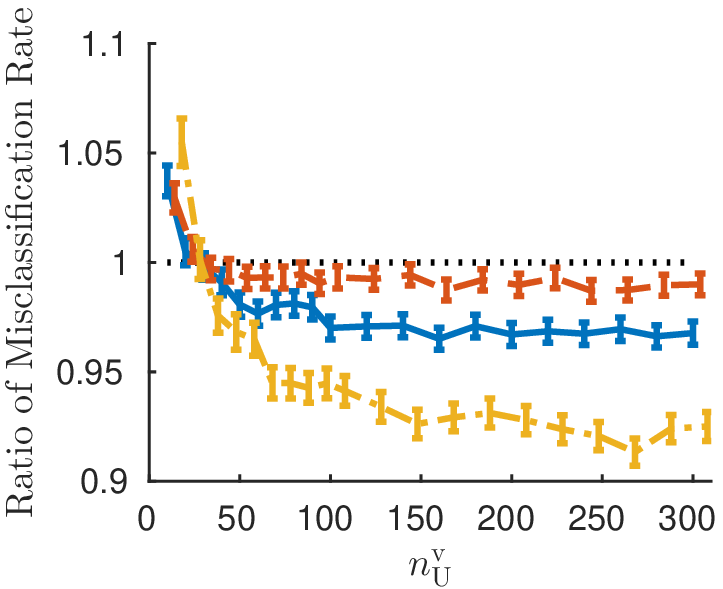}}%
 	\subfigure[Splice ($d\!=\!60$)]{%
 		\includegraphics[clip, width=.33\columnwidth]
 		{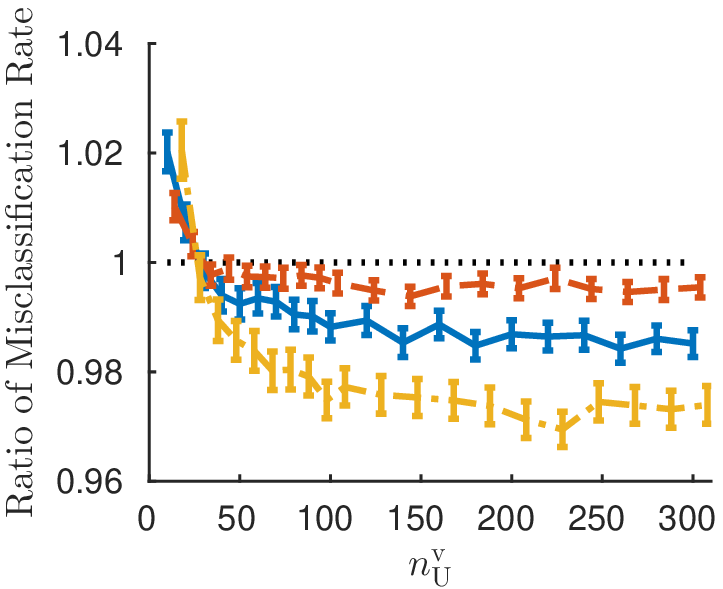}}%
	\caption{
	Average and standard error of the ratio between 
	the misclassification rate of $\widehat{g}^\PNU_\PN$ and that of $\widehat{g}^\PN_\PN$ 
	as a function of unlabeled samples over $1000$ trials. 
	In many cases, the ratio becomes less than $1$ or at worst almost $1$,
	implying that the PNU risk is a promising alternative to 
	the standard PN risk in validation
	if unlabeled data is available.
	}
	\label{app:fig:val_nu_change}
\end{figure}

Figure \ref{app:fig:var_nu_change} and Figure \ref{app:fig:val_nu_change}
respectively show the results of variance reduction and 
comparison of validation scores.  
The details of experimental setting and the interpretation of results
can be found in Section \ref{sec:exp-analyses}.

\newpage
\section{Magnified Versions of Experimental Results}
Here, we show 
magnified versions of the experimental results in Section~\ref{sec:experiments}.

\begin{table}[h]
	\centering \small
	\caption{Magnified version of Table~\ref{tab:bench}:
	Average and standard error of the misclassification rates of each
	method over $50$ trials for benchmark data sets.
    Boldface numbers denote the best and comparable methods in terms of 
	average misclassifications rate
    according to a t-test at a significance level of $5\%$.
    The bottom row gives the number of best/comparable cases of each method.    
	}		
 	\begin{tabular}{@{}c@{}r@{\,\,}r@{\,\,}r@{\,\,}r@{\!}r@{\,}r@{\!}r@{\,\,}r}
		\toprule 
		Data set & $\nl$  
 		& \multicolumn{1}{c}{PNU}  & \multicolumn{1}{c}{PUNU}
 		& \multicolumn{1}{c}{ER}    & \multicolumn{1}{c}{LapSVM}
 		& \multicolumn{1}{c}{SMIR} & \multicolumn{1}{c}{WellSVM} 
 		& \multicolumn{1}{c}{S$4$VM} \\
		\toprule

 		\cmidrule{1-9}		 	
 		\#Best/Comp.~ &  
 		& $23$ & $13$ & $11$ & $4$ & $9$ & $13$ & $7$ \\
		\bottomrule
	\end{tabular}		
\end{table}
\begin{figure}[h]
	\centering
	\includegraphics[clip,width=\columnwidth]{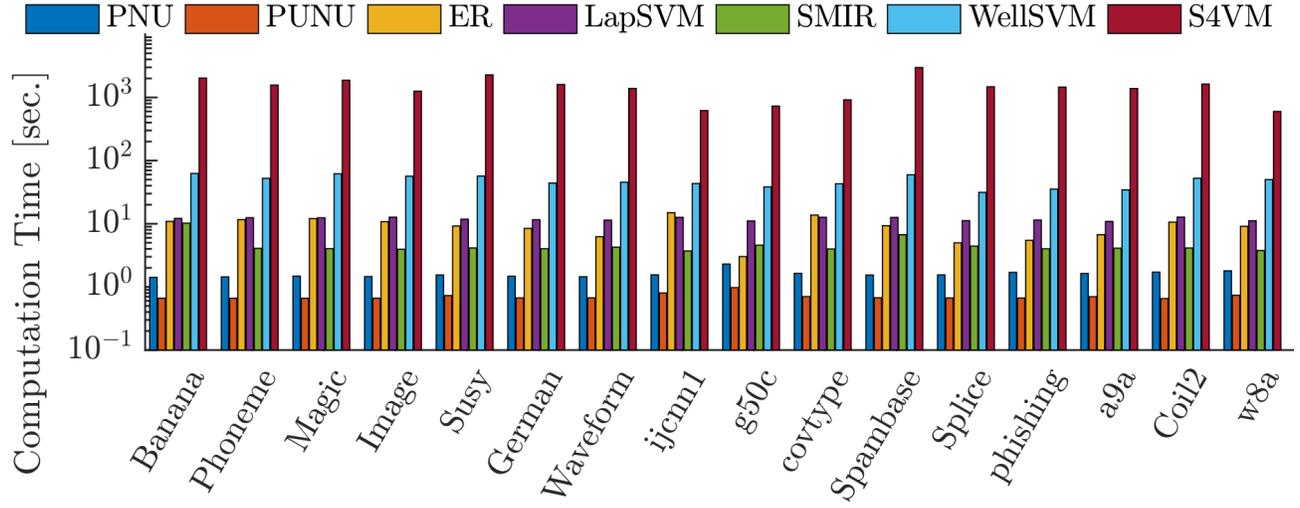}
  		\caption{Magnified version of Figure~\ref{fig:bench_time}:
  		Average computation time  over $50$ trials  
	for benchmark data sets when $\nl=50$.
	}	
\end{figure}
\begin{table}[h]		
 	\centering \small
	\caption{Magnified version of Table~\ref{tab:places_acc}:
	Average and standard error of misclassification rates 
	over $30$ trials for the Places $205$ data set.
    Boldface numbers denote the best and comparable methods 
	in terms of the average misclassification rate
    according to a t-test at a significance level of $5\%$.
	}
	\begin{tabular}{@{}l@{\,\,}r@{\,\,}r@{\,\,}r@{\,\,}r@{\,\,}r@{\,\,}r@{\,\,}r@{\,\,}r}
		\toprule 
 		Data set 
 		& \multicolumn{1}{c}{$\nun$} 
 		& \multicolumn{1}{c}{$\thetap$}
 		& \multicolumn{1}{c}{$\widehat{\theta}_\mathrm{P}$} 
 		& \multicolumn{1}{c}{PNU} 
 		& \multicolumn{1}{c}{ER} 
 		& \multicolumn{1}{c}{LapSVM} 
 		& \multicolumn{1}{c}{SMIR}
 		& \multicolumn{1}{c}{WellSVM} \\
		\midrule
\multirow{3}{*}{Arts}
& $1000$
& $0.50$ 
& $0.49$ ($0.01$) 
& $\mathbf{27.4}$ ($\mathbf{1.3}$) 
& $\mathbf{26.6}$ ($\mathbf{0.5}$) 
& $\mathbf{26.1}$ ($\mathbf{0.7}$) 
& $40.1$ ($3.9$) 
& $\mathbf{27.5}$ ($\mathbf{0.5}$) 
\\
& $5000$
& $0.50$ 
& $0.50$ ($0.01$) 
& $\mathbf{24.8}$ ($\mathbf{0.6}$) 
& $26.1$ ($0.5$) 
& $26.1$ ($0.4$) 
& $30.1$ ($1.6$) 
& N/A 
\\
& $10000$
& $0.50$ 
& $0.52$ ($0.01$) 
& $\mathbf{25.6}$ ($\mathbf{0.7}$) 
& $\mathbf{25.4}$ ($\mathbf{0.5}$) 
& $\mathbf{25.5}$ ($\mathbf{0.6}$) 
& N/A 
& N/A 
\\
\cmidrule(lr){1-9} 
\multirow{3}{*}{Deserts}
& $1000$
& $0.73$ 
& $0.67$ ($0.01$) 
& $\mathbf{13.0}$ ($\mathbf{0.5}$) 
& $15.3$ ($0.6$) 
& $16.7$ ($0.8$) 
& $17.2$ ($0.8$) 
& $18.2$ ($0.7$) 
\\
& $5000$
& $0.73$ 
& $0.67$ ($0.01$) 
& $\mathbf{13.4}$ ($\mathbf{0.4}$) 
& $\mathbf{13.3}$ ($\mathbf{0.5}$) 
& $16.6$ ($0.6$) 
& $24.4$ ($0.6$) 
& N/A 
\\
& $10000$
& $0.73$ 
& $0.68$ ($0.01$) 
& $\mathbf{13.3}$ ($\mathbf{0.5}$) 
& $\mathbf{13.7}$ ($\mathbf{0.6}$) 
& $16.8$ ($0.8$) 
& N/A 
& N/A 
\\
\cmidrule(lr){1-9}
\multirow{3}{*}{Fields}
& $1000$
& $0.65$ 
& $0.57$ ($0.01$) 
& $\mathbf{22.4}$ ($\mathbf{1.0}$) 
& $26.2$ ($1.0$) 
& $26.6$ ($1.3$) 
& $28.2$ ($1.1$) 
& $26.6$ ($0.8$) 
\\
& $5000$
& $0.65$ 
& $0.57$ ($0.01$) 
& $\mathbf{20.6}$ ($\mathbf{0.5}$) 
& $22.6$ ($0.6$) 
& $24.7$ ($0.8$) 
& $29.6$ ($1.2$) 
& N/A 
\\
& $10000$
& $0.65$ 
& $0.57$ ($0.01$) 
& $\mathbf{21.6}$ ($\mathbf{0.6}$) 
& $\mathbf{22.5}$ ($\mathbf{0.6}$) 
& $25.0$ ($0.9$) 
& N/A 
& N/A 
\\
\cmidrule(lr){1-9} 
\multirow{3}{*}{Stadiums}
& $1000$
& $0.50$ 
& $0.50$ ($0.01$) 
& $\mathbf{11.4}$ ($\mathbf{0.4}$) 
& $\mathbf{11.5}$ ($\mathbf{0.5}$) 
& $12.5$ ($0.5$) 
& $\mathbf{17.4}$ ($\mathbf{3.6}$) 
& $\mathbf{11.7}$ ($\mathbf{0.4}$) 
\\
& $5000$
& $0.50$ 
& $0.50$ ($0.01$) 
& $\mathbf{11.0}$ ($\mathbf{0.5}$) 
& $\mathbf{10.9}$ ($\mathbf{0.3}$) 
& $\mathbf{11.1}$ ($\mathbf{0.3}$) 
& $13.4$ ($0.7$) 
& N/A 
\\
& $10000$
& $0.50$ 
& $0.51$ ($0.00$) 
& $\mathbf{10.7}$ ($\mathbf{0.3}$) 
& $\mathbf{10.9}$ ($\mathbf{0.3}$) 
& $\mathbf{11.2}$ ($\mathbf{0.2}$) 
& N/A 
& N/A 
\\
\cmidrule(lr){1-9}
\multirow{3}{*}{Platforms}
& $1000$
& $0.27$ 
& $0.33$ ($0.01$) 
& $\mathbf{21.8}$ ($\mathbf{0.5}$) 
& $23.9$ ($0.6$) 
& $24.1$ ($0.5$) 
& $30.1$ ($2.3$) 
& $26.2$ ($0.8$) 
\\
& $5000$
& $0.27$ 
& $0.34$ ($0.01$) 
& $\mathbf{23.3}$ ($\mathbf{0.8}$) 
& $\mathbf{24.4}$ ($\mathbf{0.7}$) 
& $\mathbf{24.9}$ ($\mathbf{0.7}$) 
& $26.6$ ($0.3$) 
& N/A 
\\
& $10000$
& $0.27$ 
& $0.34$ ($0.01$) 
& $\mathbf{21.4}$ ($\mathbf{0.5}$) 
& $24.3$ ($0.6$) 
& $24.8$ ($0.5$) 
& N/A 
& N/A 
\\
\cmidrule(lr){1-9} 
\multirow{3}{*}{Temples}
& $1000$
& $0.55$ 
& $0.51$ ($0.01$) 
& $\mathbf{43.9}$ ($\mathbf{0.7}$) 
& $\mathbf{43.9}$ ($\mathbf{0.6}$) 
& $\mathbf{43.4}$ ($\mathbf{0.6}$) 
& $50.7$ ($1.6$) 
& $\mathbf{44.3}$ ($\mathbf{0.5}$) 
\\
& $5000$
& $0.55$ 
& $0.54$ ($0.01$) 
& $\mathbf{43.4}$ ($\mathbf{0.9}$) 
& $\mathbf{43.0}$ ($\mathbf{0.6}$) 
& $\mathbf{43.1}$ ($\mathbf{1.0}$) 
& $\mathbf{43.6}$ ($\mathbf{0.7}$) 
& N/A 
\\
& $10000$
& $0.55$ 
& $0.50$ ($0.01$) 
& $\mathbf{45.2}$ ($\mathbf{0.8}$) 
& $\mathbf{44.4}$ ($\mathbf{0.8}$) 
& $\mathbf{44.2}$ ($\mathbf{0.7}$) 
& N/A 
& N/A 
\\
		\bottomrule
	\end{tabular}
\end{table}

\end{document}